\documentclass[lettersize,journal]{IEEEtran}
\usepackage{xcolor,soul,framed} 
\usepackage{subfigure}
\colorlet{shadecolor}{yellow}
\usepackage[pdftex]{graphicx}

\usepackage{booktabs}
\graphicspath{{../pdf/}{../jpeg/}}
\DeclareGraphicsExtensions{.pdf,.jpeg,.png}
\usepackage[cmex10]{amsmath}
\usepackage{amsmath}
\usepackage{amssymb}
\usepackage{amsfonts}
\usepackage{array}
\usepackage{multirow} 
\usepackage{mdwmath}
\usepackage{mdwtab}
\usepackage{eqparbox}
\usepackage{url}
\usepackage{color}
\usepackage{algorithm}
\usepackage{algorithmic}
\usepackage[justification=centering]{caption}
\usepackage{float} 
\usepackage{dsfont}

\usepackage{amsfonts}
\usepackage{harpoon}
\usepackage[colorlinks,linkcolor=cyan,urlcolor=cyan]{hyperref}
\bstctlcite{IEEE:BSTcontrol}
\usepackage{cite}
\usepackage{makecell}
\usepackage{graphicx}
\usepackage{subcaption} 
\usepackage{adjustbox} 
\usepackage{booktabs}

\begin{document}

    \title{TT-Prune: Joint Model Pruning and Resource Allocation for Communication-efficient Time-triggered Federated Learning}
  \author{  Xinlu Zhang, Yansha Deng, \textit{Senior Member}, and Toktam Mahmoodi, \textit{Senior Member}.

  \thanks{X. Zhang, Y. Deng, and T. Mahmoodi  are with the Department of Engineering, King’s College
London, Strand, London WC2R 2LS, U.K. (e-mail: {xinlu.zhang, yansha.deng, toktam.mahmoodi}@kcl.ac.uk).

}
}


\maketitle

\begin{abstract}
Federated learning \textcolor{black}{(FL)} offers new opportunities in machine learning, particularly in addressing data privacy concerns. In contrast to conventional event-based federated learning, time-triggered federated learning \textcolor{black}{(TT-Fed)}, as a general form of both asynchronous and synchronous FL, clusters users into different tiers based on fixed time intervals. However, the FL network consists of a growing number of user devices with limited wireless bandwidth, consequently magnifying issues such as stragglers and communication overhead. In this paper, we introduce adaptive model pruning to wireless TT-Fed systems and study the problem of jointly optimizing the pruning ratio and bandwidth allocation to minimize the training loss while ensuring minimal learning latency. To answer this question, we perform convergence analysis on the gradient $l_2$-norm of the TT-Fed model based on model pruning. Based on the obtained convergence upper bound, a joint optimization problem of pruning ratio and wireless bandwidth is formulated to minimize the model training loss under a given delay threshold. Then, we derive closed-form solutions for wireless bandwidth and pruning ratio using \textcolor{black}{Karush–Kuhn–Tucker} (KKT) conditions. The simulation results show that model pruning could reduce the communication cost by 40\% while maintaining the model performance at the same level.

\end{abstract}


%
\IEEEpeerreviewmaketitle




\section{Introduction}
\IEEEPARstart{I}{n} recent years, significant advancements in sixth-generation (6G) wireless communications and machine learning have been witnessed \cite{9144301}. Portable smart devices with powerful computational capabilities and increasing utilization of various applications and services have accumulated a large amount of user data. 
However, most of this data is private, and their frequent exposure to data centers or clouds brings inevitable risks and liabilities. \textcolor{black}{The introduction of several stringent data privacy protection bills has limited the feasibility of centralized training \cite{sun2014data}.}

\textcolor{black}{The emergence of various limitations has further accelerated the advancement of wireless federated learning, establishing it as a promising alternative for privacy-preserving distributed machine learning in recent years \cite{zhang2021survey}\cite{10353003}.} 
The network does not transfer the data used for training but only uploads the model parameters from each user in each iteration. During the training process, each user is trained locally using an individually collected dataset, and then the uploaded user model, known as weights or gradients, are accepted through the server. The server aggregates the collected model parameters to generate a new global model and retransmits it to each user. These steps are repeated until the model converges \cite{mcmahan2017communication}. 
Based on the manner in which global aggregation updates are performed, FL naturally falls into two main categories: 1) In Asynchronous Federated Learning (Async FL), the global aggregation is triggered by any user's update \cite{chai2020fedat}. 2) In Synchronous (Sync) FL, the global aggregation is triggered only when updates from all users are uploaded \cite{wang2019adaptive, yang2020federated}.



Sync FL is widely studied in wireless networks considering its implementation simplicity and relatively lower communication overhead compared to Async FL. However, in practical wireless networks, bandwidth constraints and transmission unreliability can limit its communication efficiency, posing challenges for real-world deployments. Existing research has demonstrated how additional communication efficiency could be achieved. For example, the work reported in \cite{tran2019federated} optimizes FL for latency and unreliability of the wireless channel. 
However, the slowest users, known as stragglers, still affect their convergence speed and communication delay, resulting in reduced training efficiency.

By removing the need to wait for the straggler, Async FL offers a flexible solution to perform asynchronous aggregation, as soon as users become ready. This improves system efficiency by allowing devices to update the model based on their own conditions and available resources independently. \textcolor{black}{However, Async FL poses new challenges compared to the Sync FL. Its asynchronous aggregation can cause uneven training progress, inconsistent updates, and slower convergence. While it avoids synchronization delays, it may incur higher overall communication cost due to more frequent updates.} Although weighted averaging and adjusting model update frequencies can help mitigate these challenges \cite{xie2019asynchronous, xu2023asynchronous}, addressing the issue of large transmission volumes remains a key focus \cite{konevcny2016federated}. Research in this area, including wireless asynchronous federated learning, is still in its early stages.

Several studies have addressed the challenges of Async FL by directly modifying its architecture. A common approach is to cluster users based on specific criteria, such as the timing of their weight updates (straggle level), and perform inter-tier aggregation at each cluster level to improve efficiency and consistency.\cite{chai2021fedat,zhou2022time}. However, the clustering solutions often introduce additional challenges such as temporal mismatches, additional communication delays from multi-tier aggregation and \textcolor{black}{inter-cluster} coordination, and biases from incomplete local data that then in return impact model accuracy and stability. Time-triggered FL(TT-Fed) mitigates some of the aforementioned problems by conducting global aggregation at fixed time intervals \cite{zhou2022time}\cite{zhang2024joint}, structuring updates and partitioning users into tiers based on client update times. This reduces synchronization delays and communication overhead compared to fully asynchronous methods. Nevertheless, TT-Fed still suffers from biases caused by heterogeneous and incomplete local data distributions. 
Although TT-Fed significantly reduces communication overhead compared to traditional Async FL approaches, limited bandwidth resources still pose challenges for large-scale data transmission. In particular, when most devices in the system exhibit rapid updates, frequent uploads may still lead to communication bottlenecks.

Multiple studies focus on reducing the communication overhead, referred to as the communication cost, by minimizing the number of global communication rounds or reducing the amount of transmitted information. Minimizing global communication rounds requires more local updates \cite{mcmahan2017communication}, leading to a longer computation time and reduced efficiency. Reducing the amount of transmitted data could be through model compression techniques, such as sparsification \cite{jiang2022model}, quantization \cite{reisizadeh2020fedpaq}, or selective communication \cite{cho2020client}\cite{10660465}. Although these methods effectively reduce communication overhead, they introduce additional computational complexity through encoding and decoding operations, which can offset the communication savings. Model pruning offers an alternative by removing unimportant parameters, thereby reducing both communication costs and local computation time. It stands out among model compression methods for its ability to significantly lower computational complexity while generally preserving high model performance.


Based on the underlying algorithmic logic of pruning, early neural network pruning methods include second-order Taylor expansion pruning \cite{hassibi1992second}, magnitude-based pruning \cite{liu2019channel}, single-shot network pruning (SNIP) \cite{lee2018snip}, iterative pruning \cite{castellano1997iterative} and Synaptic Flow Pruning (SynFlow) \cite{tanaka2020pruning}. 
Second-order Taylor pruning evaluates weight importance using second-order model loss impact but is computationally impractical for modern DNNs. Magnitude-based pruning removes small weights but requires repeated training cycles. SNIP prunes in one step using initial gradients, trading efficiency for potential accuracy loss. Iterative pruning gradually prunes during training, balancing recovery with increased training time. SynFlow, used at initialization, assesses weight importance via simulated information flow to preserve network structure. While these methods laid the foundation for network pruning, their reliance on centralized data access and significant computational resources limits their applicability to large-scale, modern deep neural networks. In addition, several other pruning methods have been proposed for federated learning. FedDrop reduces latency by randomly dropping model parameters but requires full models during testing \cite{wen2022federated}. PruneFL is an adaptive pruning approach that performs initial pruning on selected users and further refines pruning during the FL process \cite{jiang2022model}. However, these pruning techniques do not account for multi-tier asynchronous learning scenarios. 



The incorporation of model pruning introduces new challenges for TT-Fed networks. The pruning ratio setting significantly impacts user tier assignments. A high pruning ratio greatly reduces the user's model upload and calculation delays. This reduction can cause the user's overall runtime to fall below the minimum threshold for their original tier, resulting in reassignment to a lower tier. This reassignment can alter the model status of each tier, consequently affecting overall model performance. 
Typically, simpler computational strategies based on statistical metrics (such as weight importance \cite{molchanov2019importance}) are employed to determine which parameters should be pruned. This reduces the additional computational burden required for computing the pruning rate while simultaneously reducing the model size and complexity.

Beyond reducing the size of transmitted information, it is essential to consider optimization problems within wireless networks. Unlike wired communication, the dynamic nature of wireless communication environments and the limited communication resources are the primary factors impacting learning performance. Existing studies have integrated the characteristics of wireless communication with the requirements of federated learning, proposing various targeted optimization methods. The authors of \cite{shi2020joint} proposed a joint device scheduling and resource allocation strategy for latency-constrained wireless federated learning, optimizing bandwidth and device scheduling to maximize model accuracy within a given training time budget. In \cite{chen2020joint}, the authors propose a joint optimization algorithm for wireless resource allocation and device selection to minimize the FL loss function. Notably, they analyze the convergence speed by considering the packet error rate, where the participation probability in gradient aggregation is determined by the error rate. In \cite{liu2023adaptive}, an adaptive pruning method for hierarchical federated learning is proposed, jointly optimizing pruning rates and wireless resources to reduce communication costs and latency while maintaining learning accuracy. These optimization methods have mostly been validated in synchronous federated learning, while research on wireless communication in multi-tier asynchronous federated learning remains limited.

Motivated by the above, a viable solution is to introduce model pruning and bandwidth allocation jointly for time-triggered FL in wireless networks. In this paper, we propose an optimization algorithm that combines resource allocation and pruning, balancing communication efficiency and convergence speed by jointly optimizing the pruning ratio and bandwidth allocation. 
The main contributions are summarized as follows: 
\begin{enumerate}
\item{We develop a joint optimization framework, called TT-Prune, for Time-triggered FL over resource-constrained and dynamical wireless networks by adaptively configuring uplink resource and pruning ratios. With the aim of achieving efficient communication in dynamic wireless networks, we formulate a joint model pruning and bandwidth problem.}

\item{We derive the upper bound on the $l_2$-norm of the gradients to evaluate the convergence rate. To measure the resource constraint of the wireless environment, we also derive the learning latency of the pruned time-triggered FL. Based on the analytical results, the pruning ratio and wireless resource allocation are jointly optimized to maximize the convergence rate under the given learning latency budget.}

\item{Based on the above analysis, we decouple the problem into tractable sub-problems and deployed the KKT (Karush-Kuhn-Tucker) condition. The closed-form solution of optimal pruning ratio and bandwidth allocation is formulated. Our results demonstrate that our solution can adaptively allocate bandwidth and determine the pruning ratio according to both the latency constraint and channel conditions of devices.}

\item{Compared to the TT-Fed scheme proposed in \cite{zhou2022time}, our proposed TT-prune scheme with dynamic pruning ratios and bandwidth allocation reduced the communication cost by about 40\% and achieved almost the same level of learning accuracy of no pruning TT-Fed, demonstrating the effectiveness of our proposed TT-Prune.}
\end{enumerate}


\textcolor{black}{
Although this work primarily focuses on general wireless edge learning, the proposed TT-Prune framework is also well-suited for vehicular networks (V2X), where the edge server corresponds to a roadside unit (RSU) and the clients are vehicles. Federated learning in V2X can support diverse tasks such as collaborative perception to enhance safety and traffic efficiency, shared driving policies (e.g., platooning or adaptive cruise control), and high-definition map modeling. Within these scenarios, our pruning strategy reduces communication overhead, while the resource allocation can help to accommodate the dynamic nature of V2V/V2I channels, thereby alleviating the communication challenges in vehicular networks.}

The remainder of this paper is organized as follows. Section II introduces the system model, pruning method, and latency analysis. Section III provides the convergence analysis, Section IV derives the joint optimization for pruning and bandwidth allocation, Section V presents the simulation results, and Section VI concludes the paper.

\section{System Model}

In this section, we first present a multi-tier time-triggered FL framework and then elaborate on the model pruning method. We then present the two major latencies considered in the network, followed by the problem formation associated with our proposed framework.

\subsection {TT-Fed Framework}
Consider a multi-tier wireless network consisting of an edge server, denoted as $\mathcal{K}$, a set of $\mathcal{U}$ users, denoted as $\mathcal{U}  = \{ 1,2,\ldots, U \}$. They collaborate to perform federated learning to support various intelligent applications, e.g., data analysis and image recognition. Each user has a private dataset $D_u$ and performs local model training. Each user $u \in \mathcal{U} $ possesses a local data set $D_u$, with a data volume $|D_u|$. Each data sample is represented by a set of input-output pairs $(X_{u, i}, Y_{u, i}), i\in D_u $, where $X_{u, i}$ denotes the $i$-th input data and $Y_{u, i}$ represents the corresponding ground truth for $X_{u, i}$. The entire network's data volume is represented as $D$, where $D = \sum_{u \in U} D_u$.The data is distributed throughout the network, and only a single global model parameter is transmitted within it. We use $\textbf{\textit{w}}$ to represent the network model and the loss function is denoted by  $ \ell(x_i,y_i|\textbf{\textit{w}})$ to represent the prediction error. 

We consider the time-triggered Federated Learning as the benchmark scheme \cite{zhou2022time}. In this system, participants are naturally classified into different levels according to fixed global aggregation round period $\Delta T $. Each tier includes a group of users with similar training times and can perform update operations when it is ready without waiting for updates from the slowest user in the network.
\begin{figure*}[t!]
  \centering
  \includegraphics[width=.9\textwidth]{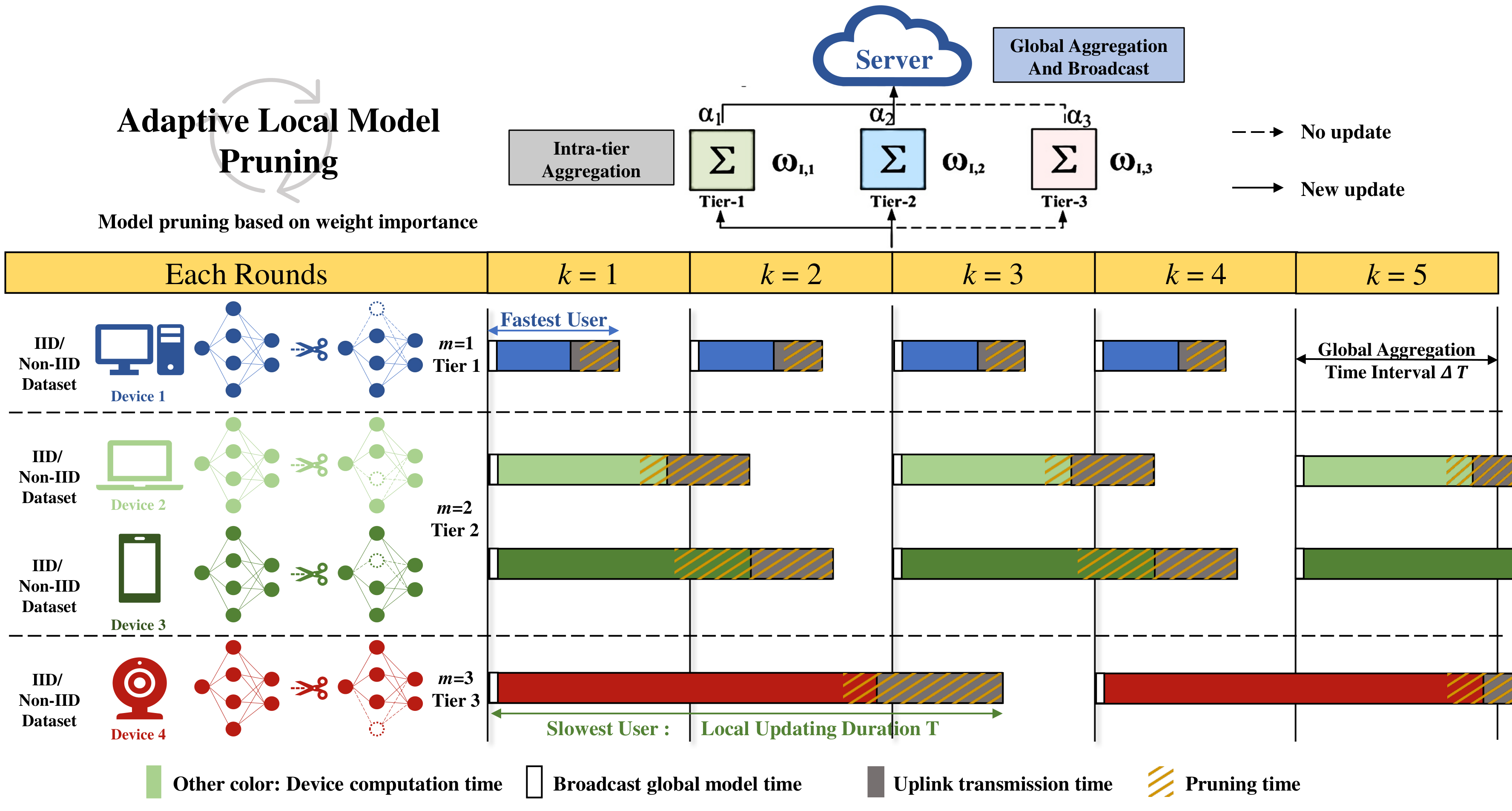} 
  \caption{The work-flow of Pruned Time-Triggered FL under the given aggregation duration $\Delta T $.} 
  \label{TTFL} 
   \vspace{-0.7cm}
\end{figure*}
Since the global aggregation is triggered by the given time $\Delta T $, all users can be automatically divided into different tiers. We assume the time for the slowest user to complete one local training round is $T$. Then, the total number of tiers we have is $M = \lceil T/\Delta T \rceil$ tiers, where $\lceil \cdot \rceil$ is the ceiling function. The first tier consists of the fastest users, and the last tier contains the slowest users. 

The specific aggregation process used in  TT-Fed is shown in Fig.  \ref{TTFL}. In a scenario with four participants, they are partitioned into three tiers based on $\Delta T$. It is important to note that users in the first tier perform local updates in every global aggregation round, users in the second tier update every two global aggregation rounds, and users in the third tier update every three global aggregation rounds. The variable $k$ represents the $k$-th round of model aggregation, and $m$ represents the $m$-th tier. To decide if the $m$-th tier joins the current aggregation, we check if $k$  \textbf{mod} $m$ is equal to zero. For example, user 4 does not participate in the 1st and 2nd global aggregation since the $ k $  \textbf{mod} $m$ $\neq 0$ ( $k=1$ or 2 and $m = 3$).

The optimization goal of TT-Fed is the same as traditional FL network, that is to minimize the empirical loss $\mathcal{L}(\mathbf{w})$ over all distributed training data $\mathcal{D}=\sum_{u \in U} \mathcal{D}_u$. The problem can be formulated as
\begin{equation}
\min _{\mathbf{w}} \mathcal{L}(\mathbf{w})=\frac{\sum_{i \in \mathcal{D}^u} \sum_{u \in \mathcal{U}} \ell\left(\mathbf{x}_{u,i}, \mathbf{y}_{u,i} \mid \mathbf{w}\right)}{|\mathcal{D}|}.
\label{eq1}
\end{equation}

The iterative learning process, which can be partitioned into four steps, is introduced as follows:
\subsubsection {Global Model Broadcast}
At the end of the ($k$-1)-th round of global aggregation, which is the beginning of the $k$-th round global aggregation, the server broadcasts the latest global model to the users who participated in the ($k$-1)-th round of aggregation. Users participating in the aggregation can be represented by \{$S_m$ $|k-1$ mod $m$ = 0, $\forall m \in M $\}. This is also because the unbroadcasted users have not completed local training, while the broadcasted users are ready to start a new round of local training. 

\subsubsection {Local Model Learning}
After receiving the global model $w_{G}^{k-1}$, the selected user $u$ will start updating the local model based on their own private dataset $D_u$. The $k$-th global aggregation may not involve selected users, as these users may require a different number of global aggregation rounds based on the tier they belong to. For example, user $u$ in the $m$-th tier needs $m$ rounds. Therefore, the $k$-th global aggregation is provided by users who belong to the $m$-th tier and are trained on the global model $w_{G}^{k-m}$. The generated local model that is ready to be uploaded can be represented as
\begin{equation}
w_{L,u}^{k} = w_G^{k-m} - \lambda \mathbb{E}\left[\nabla \ell(w_G^{k-m} ; x_{u,i}, y_{u,i})\right], \forall u \in S_m ,
\label{eq2}
\end{equation}
where $ \lambda $ is the learning rate for local model and $w_G^{k-m}$ is the global model that user\textit{u} received at the $(k-m)$-th global round. 
\subsubsection {Intra-tier Aggregation }
 Intra-tier aggregation refers to the aggregation process of local models within the same tier before global aggregation. In the $k$-th round of global aggregation, the intra-tier aggregation process of the $m$-th tier is presented as 
\begin{equation}
w_{\mathrm{I}, m}^k=\sum_{u \in \mathcal{S}_m} \frac{D_u w_{\mathrm{L}, u}^k}{\sum_{u \in \mathcal{S}_m} D_u}.
\label{eq3}
\end{equation}
\subsubsection {Global Aggregation }
The aggregation formula for the $k$-th round of global aggregation is presented as

\begin{equation}
\begin{aligned}
w_{G}^k=&\sum_{m=1}^M \ \mathds{1} \{k \bmod m=0\} \alpha_m^k w_{\mathbf{I}, m}^k \\
& +\sum_{m=1}^M(1-\ \mathds{1}\{k \bmod m=0\}) \alpha_m^k w_{G}^{k-1},
\label{eq4}
\end{aligned}
\end{equation}
where $\alpha_m^k$ is the aggregation weight of models from the $m$-th tier at the $k$-th global aggregation round, and $\sum_{m=1}^M \alpha_m^k=1$. In Eq. (\ref{eq4}), $ \ \mathds{1}\{ x \}$ is the indicator function. The first term at the right-hand side of Eq. (\ref{eq5}) is the weighted summation of uploaded models, and the second term is the latest global model $w_{\mathrm{G}}^{k-1}$ multiplied by the corresponding aggregation weight.

The occurrence of asynchronous aggregation also introduces corresponding issues \cite{zhou2022resource}. Local models based on the old global model may be detrimental to the current aggregation. Additionally, faster users contribute more to the global model and result in bias. These issues are mitigated by introducing weighting factors in the global aggregation step to control the influence of different users. To balance the computation bias towards fast tiers, we adopt a heuristic weighting scheme (as seen in Eq. (\ref{eq5}), where $\lfloor \cdot \rfloor$ denotes the floor function). Hence, the aggregation weight for models from the $m$-th tier at the $k$-th global aggregation round is given by
\begin{equation}
\alpha_m^k=\frac{\left\lfloor\frac{k}{M+1-m}\right\rfloor}{\sum_{m=1}^M\left\lfloor\frac{k}{m}\right\rfloor}.
\label{eq5}
\end{equation}

In this paper, we use TT-fed as a benchmark for pruning in the Async multi-tier FL framework. By adjusting the global aggregation time $\Delta T$, TT-Fed is transformed into a different type of FL. If $\Delta T$ is set to approach 0, each user is assigned to an independent tier. So TT-Fed transforms into Async FL, where global aggregation is triggered whenever a local update occurs. Conversely, if $\Delta T$ is set to approach infinity, it transforms into Sync FL. All users are categorized into the same tier, where all users must wait for the slowest user before aggregation. TT-Fed represents a special case of Sync and Async FL. However, when we set $\Delta T$ to be greater than the local fastest user and less than the local slowest user time, multi-tier networks appear in the entire FL system. At this time, users with similar training times are assigned to the same tier. They can be updated at any time after they are trained without waiting for the slowest user. It also alleviates the aggregation glitch phenomenon that occurs in asynchronous systems. The detailed time-triggered FL with adaptive pruning is presented in \textbf{Algorithm  \ref{alg1}.}

\subsection {Federated Learning With Model Pruning}
Nowadays, efficient and accurate applications often require large-scale neural network models. However, due to limitations in device computing power and wireless communication capacity, model compression is frequently employed to reduce the latency caused by computation and communication. Model pruning, as an effective compression technique, aims to retain model performance while removing less important weights \cite{molchanov2019importance}. Therefore, we set the model pruning ratio for the $m$-th tier in the $k$-th round of global aggregation as
\begin{equation}
\rho_{k,m}= \frac{M_{p}}{M},
\label{eq6}
\end{equation}
where \(M_{p}\) represents the number of model parameters to be pruned and $p$ denotes pruned parameters, and \(M\) represents the number of parameters of the model before pruning.

The main reason to optimize the model pruning ratio for each tier rather than for each individual user is that models within the same tier exhibit significant similarities. Due to their closely spaced update times, the impact of the pruning ratio is roughly similar for all models in the same tier. Effectively removing less important model parameters can significantly reduce the model size with only minimal computation cost. However, when the pruning ratio is high, a noticeable decline in learning accuracy may occur. The most accurate measure of weight importance is reflected by the error incurred when removing that weight. This measurement can be represented as the squared difference in prediction error between the device with the $m$-th tier's weights removed and the device with the weights retained. The larger the error, the more important the weight. Nevertheless, calculating such errors for every weight would impose a high computational cost. 

Alternatively, we select a lower-complexity method for weight importance calculation. The difference between the $u$-th device $j$-th local model weights and the updated $u$-th device $j$-th local model weights can be represented as  
\begin{equation}
 \hat{I}_{u,j} = |w_{u,j} - \hat{w}_{u,j}|,
\label{eq7}
\end{equation}
where $j$ denotes $j$-th model weight and $\hat{w}_{u,j}$ denotes the updated weight.

Note that no additional computation is required when removing unimportant weights by importance measures, as updated weights can be found in the backpropagation. This significantly reduces the processing time during the training and inference phases while also reducing the model size and easing the communication load (as seen in Eq. (\ref{eq9}), where $\left\lceil \cdot \right\rceil$ denotes the ceiling function). As a result, the total number of parameters in the entire model can be simply calculated as
\begin{equation}
W_{u, m}=\left\lceil(1-\rho_{k,m}) W_{u, m, \text { in }}\right\rceil\left\lceil(1-\rho_{k,m}) W_{u, m, \text { out }}\right\rceil,
\label{eq8}
\end{equation}
where $W_{u, m, \text {in}}$ and $W_{u,m, \text {out}}$ correspond to the number of input and output weights, respectively. We also denote $\rho_{k,m}$ as the pruning ratio of the $u$-th device in the following sections since the pruning ratio is the same at the same tier. 

\textcolor{black}{We adopt a low-complexity importance pruning method that reuses back-propagation gradients, thereby avoiding additional on-device computation. Under the per-round latency constraint of TT-Fed (Eq.~(\ref{eq24})), more complex importance estimators require extra floating-point operations (FLOPs), which increase local update time and typically need a larger pruning ratio $\rho$ to remain feasible. Our analysis shows that the convergence upper bound contains a term that grows linearly with $\rho$ (Theorem~1), indicating that smaller $\rho$ values are preferable for better convergence. Moreover, Theorems~2--4 further couple $\rho$ with bandwidth, which highlights the advantage of lightweight pruning methods in satisfying latency constraints without degrading convergence. In contrast, more sophisticated criteria such as second-order Taylor methods, SNIP, or SynFlow require computationally expensive Jacobian or Hessian calculations, or depend on centralized coordination during initialization, making them impractical for resource-constrained and privacy-preserving federated devices.
}

\subsection {Latency In Learning Process}
The TT-Fed model proposed in this paper is depicted in Fig.  \ref{TTFL}. As global iterations mainly occur during inter-tier aggregation and global aggregation, model updates are primarily occurring at the central server and local devices. We explain latency overhead by analyzing the entire global iteration process, which is divided into four main steps:

\subsubsection {Local Model Update With Pruning} When the $u$-th $m$ tier local model receives the model parameters from the $k$-th round of global aggregation, it prunes the model using a pruning mask as the pruning operation. The pruning mask of user $u$ is denoted by ${m}_{k, m}^{u}$, and the pruned parameters can be represented as

\color{black}
\begin{subequations}
\label{eq9}
\begin{equation}
m_{k,m}^{u,j} =
\begin{cases}
1, & \text{if } \hat{I}_{u,j} \in \text{Top-}(\rho \cdot d) \text{ of } \hat{I}_{u} \\
0, & \text{otherwise} ,
\end{cases}
\label{eq9a}
\end{equation}
\begin{equation}
\boldsymbol{\widetilde{w}}_{k, m}^{u} = \boldsymbol{w}_{k, m}^{u} \odot \boldsymbol{m}_{k, m}^{u},
\label{eq9b}
\end{equation}
\end{subequations}
where $d$ is the total number of weights and $\rho$ is the pruning ratio. That is, only the top $\rho \cdot d$ most important weights are preserved. Here, $\boldsymbol{m}_{k, m}^{u} = [m_{k,m}^{u,1}, m_{k,m}^{u,2}, \dots, m_{k,m}^{u,d}]$ is a binary mask vector, where each element $m_{k,m}^{u,j} \in \{0,1\}$ indicates whether the $j$-th weight is retained ($1$) or pruned ($0$). The mask values are derived based on the weight importance scores computed from Eq.~(\ref{eq7}). This operation corresponds to unstructured weight pruning implemented via a binary mask: individual weights with the highest importance are retained while the rest are zeroed.
\color{black}

Then, using gradient descent, the $u$-th device updates the pruned local model $\boldsymbol{\widetilde{w}}_{k, m}^{u}$. Given a pruning ratio $\rho_{k,m}$ of the $n$th mobile device, the number of weights after pruning is calculated as
\begin{equation}
W_{\rho_{k, m}}= (1-\rho_{k,m})W_{u}.
\label{eq10}
\end{equation}

In Eq. (\ref{eq10}), we perform weight pruning on the fully connected layer rather than the convolutional layer. This decision arises from the observation that pruning in the convolutional layer often reduces the robustness of the CNN. Since the number of weights includes convolutional layers and fully connected layers, the final number of weights for the pruned model is given by 
\begin{equation}
W_{\rho_{k, m}}= W_{u,\text{conv}}+(1-\rho_{k,m})W_{u,\text{fully}} ,
\label{eq11}
\end{equation}
where $W_{u, \text{conv}}$ represents the number of weights in the convolutional layer, and $W_{u, \text{fully}}$ denotes the number of weights in the fully connected layer.
\begin{algorithm}[t]
	\renewcommand{\algorithmicrequire}{\textbf{Input:}}
	\renewcommand{\algorithmicensure}{\textbf{Output:}}
	\caption{Time-triggered FL with joint pruning ratio and bandwidth optimization }
        \label{alg1}
\begin{algorithmic}[1]
   \REQUIRE {Local dataset $D_u$; local device $u$; one edge server;
            number of local epochs $T$; number of global rounds $K$;
            pruning mask $m_{k,m}^u$; inter-tier aggregated model $w^i_{k,m}$;
            global model $w_{G,k}$; local model $\boldsymbol{w}_{k,m}^{u}$ }

    \STATE \textbf{Initialization:} pruning-ratio vector $\rho \to 0$,
           bandwidth-allocation vector $b \to 0$
    \FOR{global round $k = 1,\dots,K$}
      \FOR{$m = 1,\dots,M$}
        \IF{$k \bmod m = 0$}
          \FOR{user $u \in S_m$}
            \STATE $\boldsymbol{w}_{k,m}^{u} \gets w_G^{k}$    
            \STATE \textbf{Compute} preliminary pruning ratio lower bound $\rho_\text{min}$ by Eq.~(\ref{eq25}) to satisfy latency constraint
            \STATE \textbf{Solve} for optimal bandwidth allocation $b^*$ via convex optimization (Eq.~\ref{eq27}) using KKT conditions
            \STATE \textbf{Compute} final pruning ratio $\rho^*$ based on allocated $b^*$ using Eq.~(\ref{eq28})
            \STATE Generate mask $\boldsymbol{m}_{k,m}^{u}$
            \STATE $\widetilde{\boldsymbol{w}}_{k,m}^{u} \gets
                   \boldsymbol{w}_{k,m}^{u} \odot \boldsymbol{m}_{k,m}^{u}$
            \FOR{local epoch $t = 1,\dots,T$}
              \STATE Update $\boldsymbol{w}_{k,m}^{u}$ as Eq.~(\ref{eq2})
            \ENDFOR
          \ENDFOR
          \STATE Update $\boldsymbol{w}_{k,m}^{i}$ on $D_u$ using Eq.~(\ref{eq14})
        \ENDIF
      \ENDFOR
      \STATE Update global model $w_{G,k}$ using Eq.~(\ref{eq4})
    \ENDFOR
    \ENSURE $k$-th global model $w_{G,k}$ for $(k{+}1)$-th aggregation
\end{algorithmic} 
\end{algorithm}

With the change in the size of the model weight, the computation latency for each user device is affected during the updating period. This is mainly because devices have different capabilities for processing data. In order to calculate the computation latency for the $u$-th local device, we first assume that the number of CPU cycles required to process a single data sample is denoted as $c_u$, and then the CPU frequency of the $u$-th user is $f_u$. Therefore, the time taken by the $u$-th user to complete local updates during the $k$-th global aggregation round is given by 
\begin{equation}
\tau_{k,u}^\text{cp}=\zeta \frac{  W_{\rho_{k,m}} c_u}{f_u},
\label{eq12}
\end{equation}
where $\zeta$ represents the number of local training epochs \cite{yang2020energy}.
\subsubsection {Local Model Uplink Transmission } Assuming the size of the model transmitted to the server is denoted by \(Z\) bits. Therefore, the achievable rate \(R_{k,u}^{up}\) and uplink communication time \(\tau_{k,u}^{cm}\) for the \(u\)-th user in the \(k\)-th global aggregation round are determined by
\begin{subequations}\label{eq13}        
\begin{align}
R_{k,u}^{\text{up}} &= b_{m,k} B \log_2\!\left(1 + \frac{p g_{k,u}}{\sigma^{2}}\right), \label{eq13a}\\[4pt]
\tau_{k,u}^{\text{cm}} &= \frac{Z}{R_{k,u}}, 
\quad \text{for all } u \in U,\;
Z = \hat{q} W_{\rho_{k,m}}. \label{eq13b}
\end{align}
\end{subequations}

The $\hat{q}$ denotes the quantization bit. The bandwidth \(b_{k,u}\) allocated to the \(u\)-th user in the \(k\)-th global aggregation round, the uplink transmission power \(p\) (assuming equal transmission power for each user), and the total noise power \(\sigma^2\). The channel gain between the $u$-th local device and central server is \(g_{k,u}\).

\subsubsection {Intra-tier Aggregation } Before the global aggregation stage, local models within the same tier are aggregated. In the $k$-th round of global aggregation, the aggregation formula for the pruning $m$-th tier is shown as
\begin{equation}
 w_{i,m}^{k,j} = \sum_{u \in N_{m}^{j}} \frac {D_{u} w_{m,u}^{k,j}}{\sum_{u \in N_{m}^{j}} D_{u}} .
\label{eq14}
\end{equation}

\subsubsection {Global Model Aggregation and Broadcast}The edge server aggregates local models from all tier users based on Eq. (\ref{eq4}) Considering that the edge server has sufficient computational resource, the computation complexity of global aggregation at the central server is low. Therefore, we will ignore the computation time at the edge server. 

After the $k$-th round of the global model aggregation process, users are selected by the edge server only from the set $\{S_m \mid k \mod m = 0, \forall m \in M\}$ to download the global model. Given the higher transmission rate in downlink communication during broadcasting of the global model, the transmission delay between them can be neglected.

\subsection {Problem Formulation}
Based on the aforementioned system model and considered latency, we formulate an optimization problem with respect to maximizing convergence speed. Maximizing convergence speed is equivalent to minimizing the $l_2$-norm of the gradient. By minimizing the $l_2$-norm of the gradient, we find an appropriate parameter update direction that allows the model to iterate faster towards the optimal point. This accelerates the convergence speed and brings the model closer to the optimal solution within a limited number of training iterations. 

Exactly, in order to maximize the convergence speed, we minimize the $l_2$-norm of the gradient. At the same time, we impose constraints on the learning process delay and make reasonable allocations for the pruning ratio and bandwidth. By considering these factors together, we aim to achieve an optimal balance between convergence speed and learning latency, ensuring efficient and effective training in wireless FL. Therefore, the optimization problem can be formulated as 
\begin{subequations}
\begin{equation}
    \min_{b_{m,k}, \rho_{k, m}} \sum_{k=1}^K \mathbb{E}\left\|\nabla F\left(w_G^k\right)\right\|^2,
    \label{eq15a}
\end{equation}
\begin{equation}
    \text{s.t.} \quad \tau_{k,u}^\text{cp} + \tau_{k,u}^\text{cm}  \leq 
 m \Delta T,
    \label{eq15b}
\end{equation}
\begin{equation}
    \sum_{m=1}^M b_{m,k}  \leq 1,
    \label{eq15c}
\end{equation}
\begin{equation}
    0 \leq b_{m,k} \leq 1,
    \label{eq15d}
\end{equation}
\begin{equation}
    \rho_{k,m} \in[0,1].
    \label{eq15e}
\end{equation}
\label{eq15}
\end{subequations}

The objective in Eq. (\ref{eq15a}) minimizes the $l_2$ norm of the gradient to maximize convergence speed. Eq. (\ref{eq15b}) ensures timely completion of the learning process for each user before the aggregation deadline. Additionally, Eq. (\ref{eq15c}) and Eq. (\ref{eq15d}) constrain the allocated bandwidth for each user within the range [0, 1]. Finally, Eq. (\ref{eq15e}) sets bounds on the model pruning ratio to maintain it between 0 and 1. These constraints collectively optimize the trade-off between convergence speed and learning latency, considering communication and resource constraints in wireless FL.

Clearly, the optimization problem in Eq. (\ref{eq15}) constitutes a mixed-integer nonlinear programming problem \cite{belotti2013mixed}, characterized by its non-convexity and impracticality in obtaining optimal solutions directly. Consequently, we decompose the original problem into several sub-problems to derive sub-optimal solutions.

\section {Convergence Analysis}
To address the issues raised in Section II, we start from the convergence analysis of Time-Triggered FL with adaptive pruning and analyze how wireless communication affects its convergence speed and accuracy. For easy implementation, we apply the same pruning ratios for each tier and same bandwidth allocation for selected users throughout the training process. Without the above two assumptions, we will revisit the pruning ratio and bandwidth allocation in Section IV.

The average $l_2$-norm is utilized to assess the convergence performance \cite{ghadimi2013stochastic, shi2019convergence}, given that neural networks are typically non-convex. Before conducting the convergence analysis, the following assumptions are made for the analysis:
\begin{enumerate}
\item{ \textbf{The L-smooth characteristic of loss function $F$}: We make the assumption that the gradient of the loss function is uniformly Lipschitz continuous with respect to $F$, i.e.:}
\begin{equation}
    \left\|\nabla F_n\left(\boldsymbol{w}_1\right)-\nabla F_n\left(\boldsymbol{w}_2\right)\right\| \leq L\left\|\boldsymbol{w}_1-\boldsymbol{w}_2\right\|,\\
     \label{eq16}
\end{equation}
where $L$ is a positive constant. $\left\|\boldsymbol{w}_1-\boldsymbol{w}_2\right\|$ is the norm of 
 $\boldsymbol{w}_1-\boldsymbol{w}_2$. This property restricts the growth rate of the function's gradient, ensuring convergence and stability when optimization algorithms are employed.
\item{ \textbf{Bounded global gradient change within $\boldsymbol{m \in  M}$ training rounds}, i.e.:}
\begin{equation}
\left(w_{\mathrm{G}}^{k-m}-w_{\mathrm{G}}^{k-1}\right)^{\top} \nabla  F\left(w_{\mathrm{G}}^{k-1}\right) \leq  \delta\left\|\nabla  F\left(w_{\mathrm{G}}^{k-1}\right)\right\|^2 ,
\label{eq17}
\end{equation}
\begin{equation}
     \left\|w_{\mathrm{G}}^{k-m}-w_{\mathrm{G}}^{k-1}\right\| \leq  \varepsilon ,
     \label{eq18}
\end{equation}
where $\delta$ and $\varepsilon$ are positive constants. Eq. (\ref{eq17}) can be readily fulfilled by leveraging Eq. (\ref{eq18}) and the Cauchy-Schwarz inequality. This is because the left-hand side of the inequality represents the inner product of the vectors $\left(w_{\mathrm{G}}^{k-m}-w_{\mathrm{G}}^{k-1}\right)$ and $\nabla F\left(w_{\mathrm{G}}^{k-1}\right)$. Each user updates the model using only its local data rather than the global dataset. As a result, each user's contribution is local, leading to bounded changes in the global gradient \cite{dinh2020federated}.
\item{ \textbf{Bounded local gradient change within $\boldsymbol{m\in M}$ training rounds}, i.e.:}
\begin{subequations}\label{eq19}          
\begin{gather}
\|\nabla f(w_{\mathrm{G}}^{k-m})\| \le \beta\|\nabla f(w_{\mathrm{G}}^{k-1})\| 
  \tag{\color{black}{19a}}\label{eq19a}\\[4pt]
\|\nabla f(w_{\mathrm{G}}^{k-1})-\nabla f(w_{\mathrm{G}}^{k-m})\| \le \phi ,
  \tag{\color{black}{19b}}\label{eq19b}
\end{gather}
\end{subequations}

where $\beta$ and $\phi$ are both positive constants. Comparable assumptions are frequently encountered in the convergence analysis of Async FL, as documented in \cite{chai2020fedat} and \cite{xie2019asynchronous}.
\item{ \textbf{Pruning-induced Noise in convergence}, i.e.:}
\begin{equation}
    \mathbb{E}\left\|\boldsymbol{w}_{k, n}^{q, e}-\boldsymbol{w}_{k, n}^{q, e} \odot \boldsymbol{m}_{k, n}^{q, e}\right\|^2 \leq \rho_{n, e} D^2,
    \label{eq20}
\end{equation}
where \textit{D} is a positive constant. To account for the impact of pruning on neural network convergence analysis \cite{stich2018sparsified}, we examine the effect of pruning-induced noise. The model error of the $u$-th device in $m$-tier, subject to the pruning ratio $\rho_{k, m}$, is constrained by Eq. (\ref{eq20}).
\end{enumerate}

These assumptions, which are commonly met by widely used loss functions such as the cross-entropy function, lay the foundation for deriving the convergence rate of TT-FL with network pruning, as demonstrated in the following theorem.

\textbf{Theorem 1:} When the above assumptions hold and $\frac{1}{4 L} > \delta$, the upper bound on the average $l_2$-norm of the gradient after $S$ iterations in TT-Fed using adaptive pruning can be derived as
\begin{equation}
\begin{aligned}
&\frac{\xi}{4L} \sum_{k=1}^K \mathbb{E}\left\|\nabla F\left(w_G^k\right)\right\|^2 \leq \\
&\frac{\mathbb{E}\left[F\left(w^0\right)\right]-
\left[F\left(w^*\right)\right]}{1-4L\delta} +\xi \frac{3LD^2}{(1-4L\delta)M} \sum_{k=1}^K \sum_{m=1}^M \frac{S_m^2}{D_m^2} \rho_{k, m}\\
& + \frac{K \xi L \varepsilon^2}{1-4L\delta} +\frac{\xi K}{(1-4L\delta)2M} (3L  \Omega_1 +\frac{M-1}{LD^2}  \Omega_2)   .
\label{eq21}
\end{aligned}
\end{equation}

In Eq. (\ref{eq21}), $K$ is the number of global communication rounds, $M$ is the total number of model tiers, $D$ represents the total number of data samples used in the training process, $L$ is the positive constants in assumption 1, and $S_m$ denotes the number of users in $m$-th tier. By denoting $\xi$ as the parameter at the median value theorem, we have 
\begin{equation}
    \Omega_1 = \sum_{m=1}^M \frac{S_m^2}{D_m^2} \varepsilon^2  \quad \text{and} \quad
    \Omega_2 = \sum_{m=1}^M \sum_{j=1, j \neq m}^M S_j^2 \phi^2  .
\label{eq22}
\end{equation}

\textit{Proof}: Please refer to Appendix A.

\textit{Remark 1}: From Theorem 1, we obtain the upper bound of the average $l_2$-norm of the gradient, which represents the convergence speed of the model after incorporating model pruning. When the right side of the equation is smaller, the model converges faster. Additionally, the relationship between the model pruning ratio and convergence speed can be observed: as the model pruning ratio increases, the right side of the equation increases, leading to a decrease in convergence speed. Therefore, to optimize the model's performance, we need to minimize the right side of the equation.

As a result, we now formulate the final optimization problem aimed at minimizing the specific upper bound described in Theorem 1. The problem Eq. (\ref{eq15a}) can be expressed as
\begin{equation}
\min \left\{ \xi \frac{3LD^2}{(1-4L\delta)M} \sum_{k=1}^K \sum_{m=1}^M \frac{S_m^2}{D_m^2}  \rho_{k, m}\right\}.
\label{eq23}
\end{equation}

\section {Optimal Solution}
In this section, we present an optimal solution to address the problem defined in Section II, with the objective of determining the optimal pruning ratio and bandwidth allocation. Our goal is to decompose the complex optimization problem into sub-problems.

\subsection{Optimal Pruning Ratio Analysis}
The time for a user to complete a global aggregation includes downlink distribution time, local calculation time, and uplink communication time. Thus, the overall time constraint on each tier should satisfy
\begin{equation}
\begin{aligned}
 \underbrace{T_{\text {downlink }}+ T_{\text {computation }}} + T_{\text {uplink }}  &\leq  \\
T^{\text{cmp}}_{m, k, u} \quad  + \quad T^{\text {uplink }}_{m, k, u} & = \\
 \left(1-\rho_{k,m}\right) W_{k,m}^u\left(\zeta \frac{c_u}{f_u}+\frac{\hat{q}}{R_{m, u}^k}\right) &\leq m \Delta T .
\end{aligned}
\label{eq24}
\end{equation}



As we mentioned, downlink communication can be ignored because of its capacity for communication. The original problem is non-convex because the optimization of both the pruning ratio and bandwidth allocation is performed simultaneously. Hence, we decompose the problem to derive the optimal solution. 

Once a constant value is assigned to the bandwidth, the problem described in Eq. (\ref{eq21}) can be viewed as a linear programming problem. Since the optimization solution is set the same for each tier, $u$ here represents the average value of users in each tier. Subsequently, we derive the optimal solution for the pruning ratio based on the theorem presented below.

\textbf{Theorem 2:} Under the given bandwidth allocation, the optimal pruning ratio of the device in the $m$-tier associated with $k$-th global communication round should satisfy
\begin{equation}
     \rho^*_{k,m} \geqslant \left(1-\frac{m \Delta T - W_{u, \text{conv}} \left(\xi \frac{ c_u}{f_u}+\frac{\hat{q}}{R_{m,u}^{\text{k}}}\right)  }{(\xi \frac{ c_u}{f_u}+\frac{\hat{q}}{R_{m,u}^{\text{k}}})W_{u, \text{fully}}}\right)^+ ,
     \label{eq25}
\end{equation}
where the function \((x)^+\) means \(max(x,0)^+\).

\textit{Proof}: Please refer to Appendix B.

\textit{Remark 2}: From theorem 2, the optimal pruning ratio increases/decreases with the number of weights in the model, the transmission rate, and the local computational power. In addition, when the user's computational power is held constant, the pruning ratio of the model decreases as the bandwidth increases. Conversely, when the bandwidth decreases, the pruning ratio increases. This also means that when the bandwidth increases, the channel has sufficient capacity to transmit larger model sizes to improve the convergence rate.

\subsection{Optimal Bandwidth Allocation}
According to the optimal pruning ratio and the upper bound of the average $l_2$-norm of the gradient provided by the aforementioned theorem, the original problem can be rewritten as (subject to the constraints in Eq. (\ref{eq15c}) and (\ref{eq15d}))
\begin{equation}
      \min_{b^*,\rho^*}  \boldsymbol{\Delta} \sum_{k=1}^K \sum_{m=1}^M \frac{S_m^2}{D_m^2}\left(1-\frac{m \Delta T - W_{u, \text{conv}} \left(\xi \frac{ c_u}{f_u}+\frac{\hat{q}}{R_{m,u}^{\text{k}}}\right)  }{(\xi \frac{ c_u}{f_u}+\frac{\hat{q}}{R_{m,u}^{\text{k}}})W_{u, \text{fully}}}\right),
      \label{eq26}
\end{equation}
where $\boldsymbol{\Delta} =  \frac{3LD^2}{(1-4L\delta)M} $. In order to derive the optimal solution for bandwidth, we need to minimize Eq. (\ref{eq26}). 

First, by performing multiple derivative operations on Eq. (\ref{eq26}), we need to first prove that the problem is convex. 

\textbf{Lemma 1: The problem Eq. (\ref{eq26}) is convex. }

For details on the specific calculation process, please refer to Appendix C.

Subsequently, by using the Lagrange multiplier method \cite{bomze2010interior}, the optimal wireless bandwidth is given in the following theorem.

\textbf{Theorem 3:} Setting $\lambda^*$ as the optimal Lagrange multiplier, to achieve optimal FL performance, the optimal bandwidth allocated to the $u$-th device in the $m$-th tier should satisfy
\begin{equation}
b_{m,k}^*=\frac{f_u\left(\sqrt{\frac{\boldsymbol{\Delta}S_m^2 m\Delta T\hat{q}W_{u, \text{fully}} B  \log _2\left(1+\frac{p g_{k,u}}{\sigma^2}\right)}{D_m^2 \lambda^*}}-\hat{q}W_{u, \text{fully}}\right)}{BW_{u, \text{fully}}\xi c_u  \log _2\left(1+\frac{p g_{k,u}}{\sigma^2}\right)}.
\label{eq27}
\end{equation}

To dynamically adapt to the changing wireless network environment over time, the optimized algorithm needs to be executed once during each global aggregation.

\begin{figure*}[t!]
\centering
\subfigure[Test accuracy for four different schemes on two-Tier TT-Fed]{
\includegraphics[width=7.4cm]{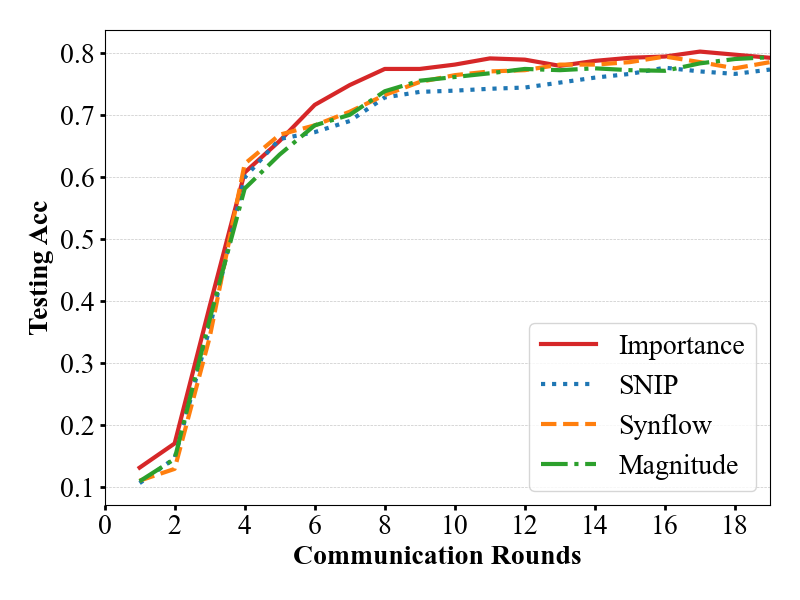}
}
\quad \quad \quad\quad \quad
\subfigure[Communication cost for IID FMNIST]{
\includegraphics[width=7.4cm]{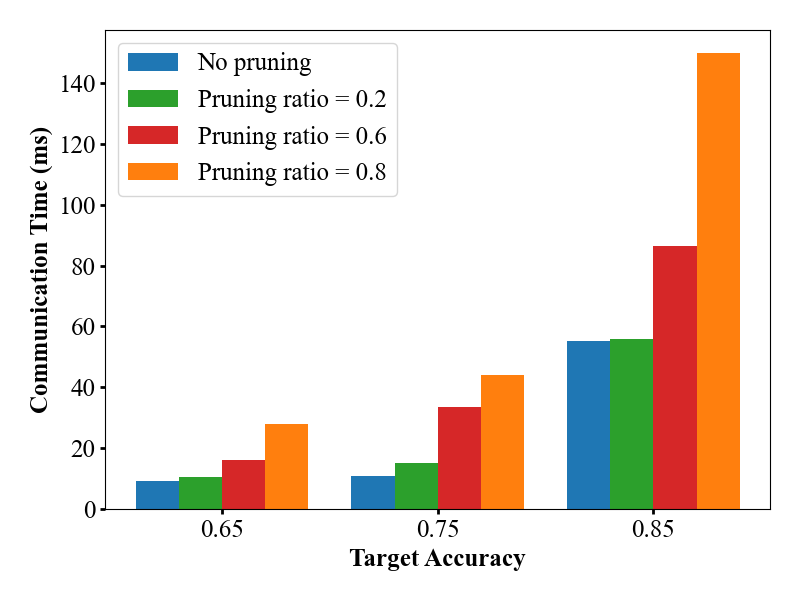}
}
\quad
\caption{Test accuracy for four different schemes on two-Tier TT-Fed}
\label{fig2}
\vspace{-0.7cm}
\end{figure*}

\textit{Proof}: Please refer to Appendix D.

\textit{Remark 3}: Similarly, it can be observed that, given the computational capacity of the user devices, according to Theorem 3, the bandwidth allocation decreases with the improvement of channel conditions. More wireless resources are allocated to user devices with poorer channel conditions to ensure transmission latency.

\textbf{Theorem 4:} Furthermore, based on Theorem 3, we can rewrite the final optimized pruning ratio, which can be expressed by optimal bandwidth allocation as
\begin{equation}
\begin{aligned}
&\rho^*_{m, k} = \\
&1-\frac{b_{m,k}^*B\log _2\left(1+\frac{p g_{k,u}}{\sigma^2}\right)(m \Delta T - W_{u, \text{conv}}\xi \frac{ c_u}{f_u})-\hat{q}W_{u, \text{conv}} }{b_{m,k}^*B\log _2\left(1+\frac{p g_{k,u}}{\sigma^2}\right)\xi \frac{ c_u}{f_u}+\hat{q}W_{u, \text{fully}}}.
\label{eq28}
\end{aligned}
\end{equation}

In addition to the proposed optimization strategies TT-Prune, this paper also presents our proposed TT-prune and three baselines under TT-FL framework as follows:
\begin{enumerate}
    \item \textbf{TT-Prune}: Both the pruning ratio and bandwidth allocation are dynamically optimized according to Eq. (\ref{eq27}).\color{black}
    \item \textbf{Equal Resource}: The pruning ratio is optimized according to Eq. (\ref{eq27}) while the bandwidth is equally allocated to all users. 
    \item \textbf{FedAvg}: The standard federated learning algorithm where all clients train locally and the server aggregates their models through weighted averaging. \color{black}
    \item \textbf{No Pruning}: The bandwidth is equally allocated to all users, and model pruning is not implemented.
\end{enumerate}

\subsection{Computational Complexity Analysis}
The proposed optimization algorithm shows that computational complexity mainly arises from the calculations of the optimal pruning ratio and the wireless resource allocation. Based on Eq. (\ref{eq28}), $\rho^*_{m, k}$ can be derived from $ b^*_{m, k}$ and Each tier is independent of the others. Therefore, the computational complexity is primarily determined by Eq.  (\ref{eq27}). As a result, the main computational complexity for each tier, derived from the interior-point method \cite{bomze2010interior} and bandwidth allocation, is, 
$
O(K^{3.5}L\log\left(\frac{1}{\epsilon}\right)),
$
where \( L \) is a Lipschitz constant and \( \epsilon \) is the accuracy requirement.
\subsection{Tier Reassignment Robustness Analysis}
\textcolor{black}{Tier reassignment is rare in our framework because the objective discourage excessive pruning, the per-tier latency constraint determines near-minimal pruning levels, and joint resource allocation stabilizes client tiers. Any occasional reassignments have minimal impact due to our aggregation weighting scheme that reduces inter-tier bias. To verify this, we conduct experiments with deliberate tier reassignments. Results confirm that individual client reassignments have negligible impact on model performance.}

\section {Numerical Results}

In this section, we evaluate the effectiveness of our proposed TT-Prune under TT-Fed framework over wireless networks and compare it with the original TT-Fed framework proposed in \cite{zhou2022time} as well as three baseline frameworks.

In the simulations, we consider a circular network area having a radius $ r = 500 m $ with one BS, inside which $U$ = 20 users are uniformly distributed. Each user trains a traditional CNN model for image classification task using the cross-entry loss function. To validate the effectiveness of the proposed algorithm, we employ the MNIST, Fashion MNIST (FMNIST), and \textcolor{black}{CIFAR-10} dataset \cite{lecun1998gradient}, each consisting of 50,000 training images and 10,000 validation images. We use both IID(independent and identically distributed) and non-IID datasets. In IID, the training data is shuffled to ensure that the data obtained for each user is IID. In non-IID, the amount of data obtained by each user is the same, but there is a certain deviation between each user. Users only need to transfer the parameters of the model to the server. We are using a CNN network. When training these two data sets, the sizes of the input layer and output layer are the same, $1 \times 28 \times 28$ and $10$, respectively. The remaining convolutional layer, pooling layer and fully connected layer each have two layers. \textcolor{black}{For CIFAR, we adopt ResNet-18, a residual architecture consisting of four stages of residual blocks with shortcut connections, enabling deeper network training and serving as a widely used benchmark on CIFAR-10.} Other key simulation parameters are listed in Table \ref{T1}.

\begin{table}[htbp]
  \centering
  \begin{tabular}[l]{@{}lc}
  \toprule
       \textbf{\emph{Parameter type}}  & \textbf{\emph{Value}}\\ 
       \hline 
       \hline
      Transmission power of device &  28 dBm \\
       CPU frequency of device & 5 Ghz  \\
       AWGN noise power &   -174 dBm/Hz \\ 
      Quantization bit, $\hat{q}$ &  64 \\ 
        Number of devices &  20  \\ 
       Number of local epochs &  5 \\ 
        Number of global communication rounds &    10 \\ 
       Learning rate  &  0.0015 \\ 
        Bandwidth of Channel &  20MHz \\ 
        Batch size   &  64 \\ 
        $\Delta$T & 0.7 T \\

    \bottomrule
  \end{tabular}
   \caption{Simulation Setup}
   \label{T1}
    \vspace{-1cm}
\end{table}

\subsection{Impact of Pruning method and Model Pruning Ratio}
\color{black}
We first analyze the impact of four classical pruning methods on the TT-Fed model, and then select importance-driven pruning to further examine its effects under different pruning ratios and IID datasets. For TT-Fed, we set $\Delta T = 0.6T$, enabling the system to naturally transition into a two-tier framework.
\begin{figure}[t!]
\centering
\includegraphics[width=7.4cm]{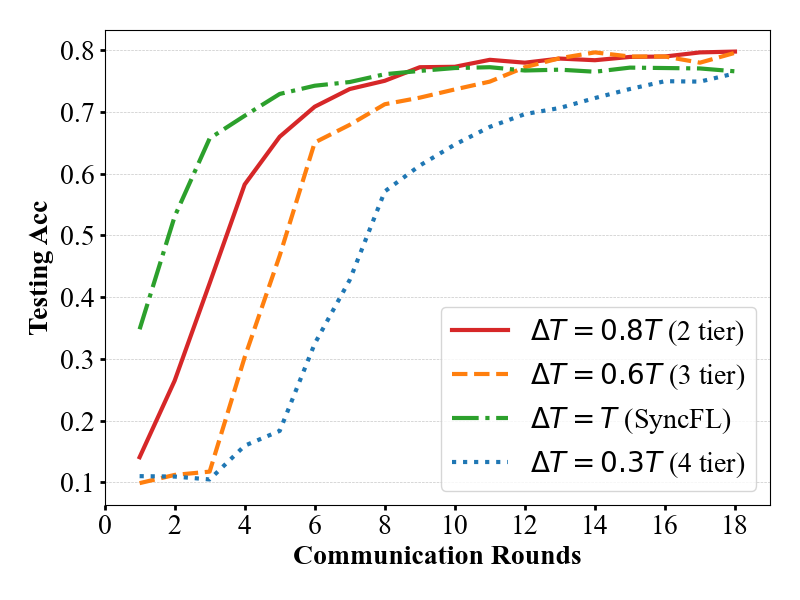}
\caption{Test accuracy of TT-Fed under different time constraints on CIFAR}
\label{fig3}
\vspace{-0.37cm}
\end{figure}

\begin{figure*}[h]
\centering
\subfigure[\textcolor{black}{IID CIFAR}]{
\includegraphics[width=7.4cm]{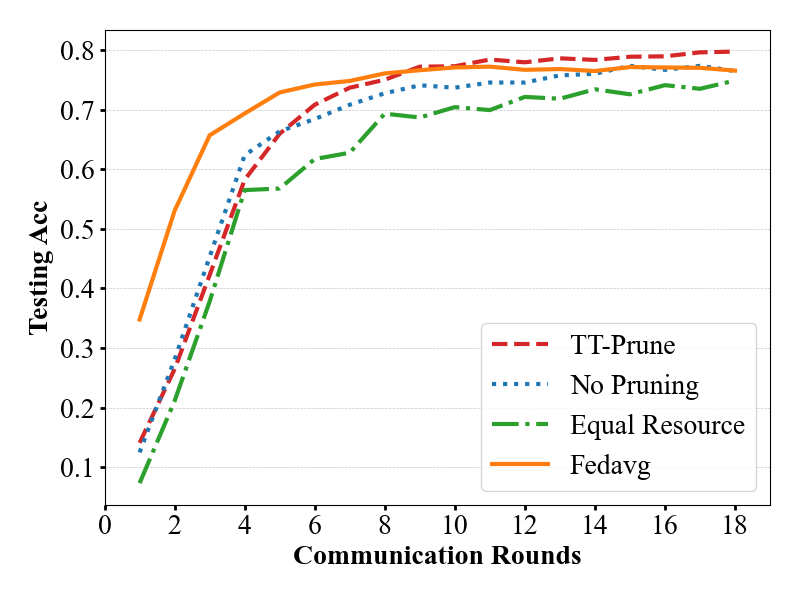}
}
\quad \quad \quad\quad\quad
\subfigure[Non-IID FMNIST]{
\includegraphics[width=7.4cm]{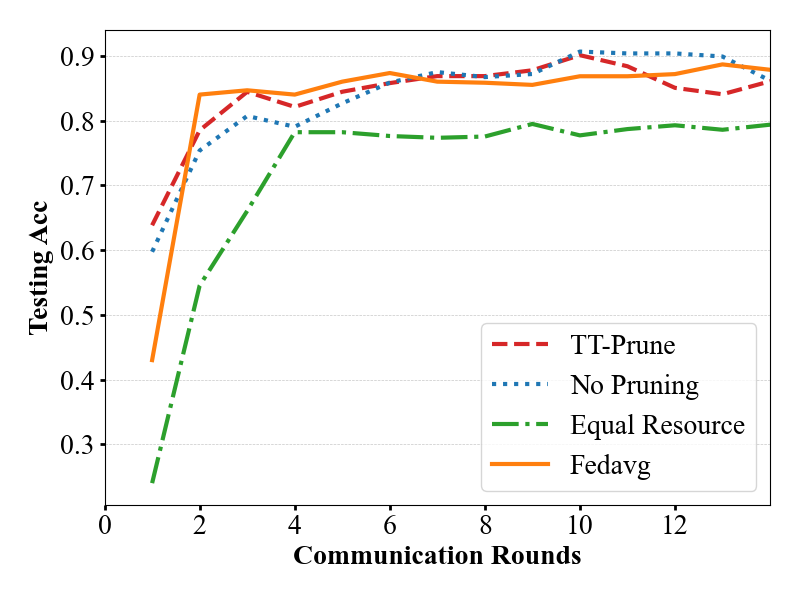}
}
\quad
\caption{Model performance for four different schemes on two-Tier TT-Fed}
\label{fig4}
\vspace{-0.7cm}
\end{figure*}

Fig.~\ref{fig2}(a) shows the test accuracy of TT-Fed on the CIFAR IID dataset using four pruning methods: Importance, SNIP, SynFlow, and Magnitude pruning. The results indicate that all four methods achieve very similar accuracy across communication rounds, while our importance-driven rule attains comparable or slightly higher accuracy during several training phases, particularly in the early and middle rounds. This shows that the chosen pruning rule is competitive in terms of accuracy. The similarity in pruning effects is mainly due to the fact that only neurons in the fully connected layers are pruned, which does not substantially affect the overall model performance. In addition, the pruning ratio is limited by the per-layer time budget and therefore cannot be very large, which further reduces the performance differences among the methods. Moreover, under comparable pruning effects, the proposed importance-driven pruning requires a lower computational cost, enabling the model to achieve better performance within a shorter training time. This aligns with our design objective, and hence we adopt importance-driven pruning in subsequent experiments to achieve optimization goals.

Fig. \ref{fig2} (b) illustrates the impact of three pruning ratios on communication cost under the FMNIST dataset with pruning ratios of 0.2, 0.6, and 0.8. It is evident that higher pruning ratios result in longer training times to reach the target accuracy.This phenomenon occurs because a higher pruning ratio leads to more weights being pruned, resulting in increased model aggregation error. Consequently, training the learning model requires more iterations and time. We also observe that the communication time at a pruning ratio of 0.2 is close to that of the unpruned case. This means that lower pruning ratios (around 0.2) have the potential to reduce communication costs without significantly compromising accuracy. Moreover, as the pruning ratio increases, the accuracy tends to degrade more rapidly. This effect can be attributed to the removal of a larger number of parameters at higher pruning ratios, which reduces the model’s capacity to capture complex data patterns, thereby accelerating accuracy degradation. These observations are consistent with the analysis highlighted in  \textit{Remark 1}, where increasing the pruning ratio, while keeping all other parameters fixed, slows down the convergence of the model.

\color{black}

\subsection{Optimal Pruning over Wireless Network}
This subsection evaluates our proposed TT-Prune's performance by comparing its test accuracy with those of all other considered optimization approaches.
\subsubsection{Impact of different global aggregations times}

In Fig. \ref{fig3}, we plot the impact of the proposed TT-Prune on test accuracy using \textcolor{black}{CIFAR} under various delay budgets. We denote the delay budget using $\Delta T$ (denote the global aggregation time), where we consider $\Delta T = T$, $\Delta T = 0.8T$, $\Delta T = 0.6T$, and $\Delta T = 0.6T$. Given the configuration of TT-Fed, the network naturally segregates into tier one to three. It can be seen that the choice of learning delay budget significantly influences convergence speed. As the delay constraint increases (as $\Delta T$ decreases), the model converges more slowly, requiring more iterations to attain convergence under higher delay constraints. Due to the need to meet larger delay constraints, larger pruning ratios are chosen to satisfy the latency requirements. However, this comes at the cost of sacrificing the model's learning performance, leading to a decrease in convergence speed. On the contrary, with a sufficient delay budget, we can achieve a higher convergence rate with a smaller pruning ratio.

To better demonstrate the advantages of the proposed TT-Prune on the model, we set the two-tier TT-Fed for pruning.

When employing the TT-Prune scheme, if the model only conducts first-tier device aggregation, it changes to a single-tier aggregation state. It cannot benefit from multi-tier optimization strategies in this state. Consequently, optimization is performed individually for each user. To demonstrate the performance of TT-Prune, we compare it with FedAvg, Equal Resource, and No Pruning, as mentioned in Section IV, with both Fig. \ref{fig4} and Fig. \ref{fig5} evaluating and comparing our proposed TT-Prune with these three schemes.

\subsubsection{IID and Non-IID Dataset Performance}
\color{black}
We selected CIFAR (IID) and FMNIST (non-IID) for for model training and testing. The CIFAR dataset is used to showcase the adaptability of the proposed model to more complex architectures, whereas FMNIST is included to demonstrate its robustness in handling smaller and more sensitive datasets.

Fig.~\ref{fig4} presents the testing accuracy on CIFAR under four schemes: FedAvg, No Pruning, TT-Prune, and Equal Resource. FedAvg shows slightly faster convergence at the early stage, whereas TT-Prune rapidly narrows the gap and achieves comparable or even slightly higher accuracy than both FedAvg and No-Pruning over communication rounds. Equal Resource performs slightly worse than the others, particularly in the early and middle stages. Importantly, TT-Prune achieves accuracy comparable to the strongest baselines. Furthermore, the optimized pruning and bandwidth allocation lead to lower communication latency per round, effectively achieving our objective of reducing total training time. These results validate the effectiveness of TT-Prune under the CIFAR setting and motivate subsequent experiments on additional datasets and model architectures.

Additionally, the figure shows that the CIFAR curves exhibit smoother trends with smaller performance gaps among methods. The capacity of ResNet and the diversity of the dataset help maintain stable learning under tiered updates. In contrast, the FMNIST curves under non-IID settings display more pronounced fluctuations. The data heterogeneity across clients reduces update consistency, making the system more sensitive to variations in pruning and bandwidth allocation, particularly during the middle stages of training. This highlights an interesting tradeoff: while FMNIST appears simpler in terms of visual complexity, the combination of its non-IID partitioning and the more limited capacity of the CNN model amplifies sensitivity to algorithmic design choices. Conversely, although CIFAR presents more complex visual patterns, ResNet's higher model capacity effectively absorbs this complexity, yielding more stable convergence behavior.

\begin{figure}[htbp]
\centering
\includegraphics[width=7.4cm]{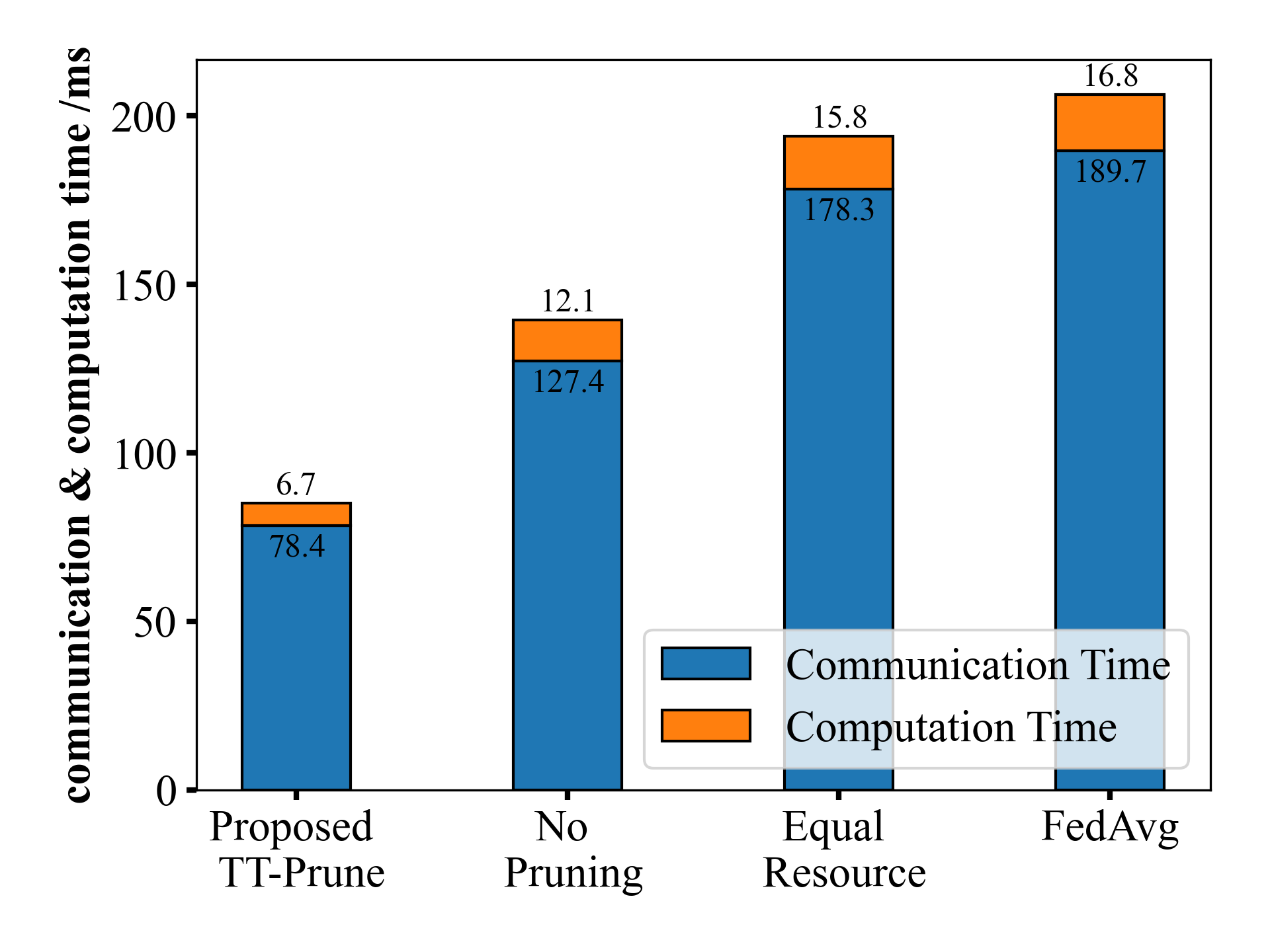}
\caption{ Performance required for TT-Prune and other schemes to achieve 80\% accuracy in Non-IID FMNIST dataset }
\label{fig5}
\vspace{-0.4cm}
\end{figure}

Fig. \ref{fig5} plots the time required to reach 80\% accuracy for TT-Prune, No Pruning, Equal Resource and FedAvg. TT-Prune achieves the best performance, reducing total training time by approximately 40\% compared to the No Pruning baseline. This significant improvement comes from adaptively pruning less important weights based on channel conditions, which reduces communication overhead while preserving model quality. In contrast, the other three schemes all require considerably more time to reach the target accuracy. Among them, No Pruning performs relatively better due to its multi-layer aggregation that improves transmission efficiency. FedAvg takes even longer as it transmits full models without any compression. Equal Resource requires the most training time, since its aggressive pruning strategy removes important weights and necessitates additional communication rounds to compensate for the degraded model quality.
\color{black}

\begin{figure}[t!]
\centering
\includegraphics[width=7.4cm]{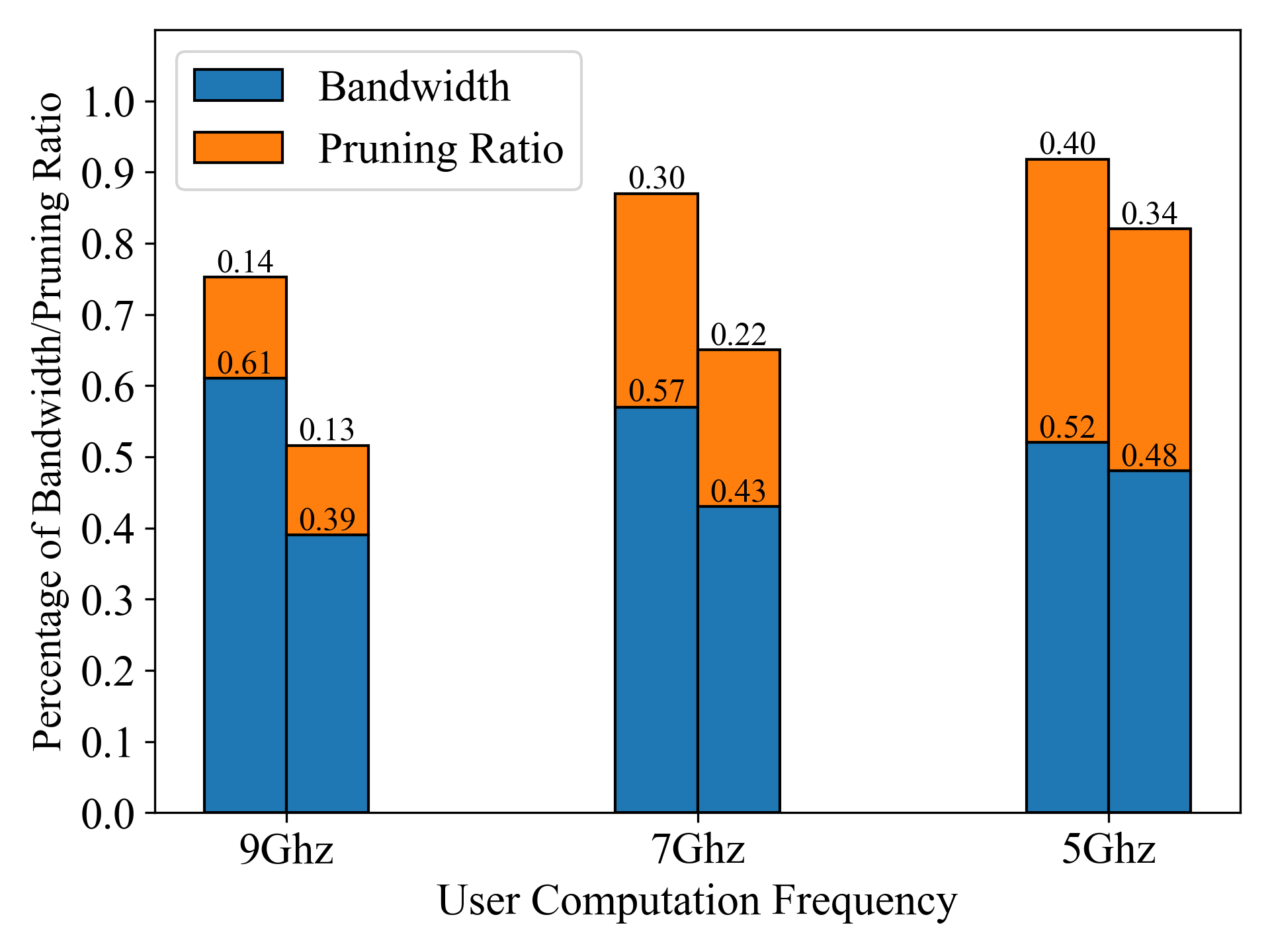}
\caption{ Relationship among pruning ratio, wireless bandwidth allocation, and computation capability in TT-Prune.}
\label{fig6}
\vspace{-0.7cm}
\end{figure}

Fig. \ref{fig6} presents the relationship between model pruning ratio and bandwidth allocation across devices with varying computational capabilities. 
Fig. \ref{fig6} is obtained from our optimization of each tier. The left side is tier 1, and the right side is tier 2. The number of users in the two tiers is different, with a ratio of 6:4. This observation further confirms the inference drawn in \textit{Remark2}. From the observations in the figures, it can be seen that under the same computing power, when more bandwidth is given to the local device, a smaller pruning ratio can be adopted to ensure a higher convergence speed. Moreover, for local devices with higher computing capabilities, allocating more wireless resource and opting for a smaller pruning ratio help mitigate computing and communication delays, ultimately enhancing convergence speed.

\section{Conclusion}
In this paper, we proposed a joint optimization framework for model pruning and bandwidth allocation in wireless Time-triggered federated learning to enhance communication and learning efficiency. We proposed an adaptive model pruning to reduce model size based on network conditions. We first conducted the convergence analysis of an upper bound on the $l_2$-norm of gradients in pruned TT-Fed to evaluate its performance. Subsequently, we utilized Karush-Kuhn-Tucker conditions to jointly optimize the pruning ratio and wireless resource allocation under latency and bandwidth constraints. Simulation results shown that our proposed TT-Prune reduces the total latency by approximately 40\% compared to models without pruning and single-optimized schemes while maintaining the same level of learning accuracy. This confirms the effectiveness of our proposed TT-Prune.
It marks the first consideration of model pruning in wireless TT-Fed systems. Our future work will focus on enhancing our approach by integrating additional optimization techniques, such as air computing and model compression.



%

\appendices
\section{Proof of the Theorem 1 }


Our goal is to analyze the upper bound of $\mathbb{E}\left\|\nabla F\left(\omega_G^k\right)\right\|^2 $ with respect to the pruning ratio $\rho$. 
At first, we have \\
{\small\begin{equation}
\begin{aligned}
&F(w_G^{k+1}) - F(w_G^k)= \\
&F\left[\sum_{m=1}^M\left(1-\mathds{1}\{(k+1) \bmod m=0\} \right) \alpha_m^{k+1} w_G^{k} \right. \\
&+ \left.\sum_{m=1}^M\mathds{1}\{(k+1) \bmod m=0\} \alpha_m^{k+1} w_{I,m}^{k+1}\right] - F(w_G^k).
\end{aligned}
\end{equation}}

Because of the convexity of $F$ and Jensen's inequality, upon rearranging the inequality, we get 
{\small
\begin{equation}
\begin{aligned}
&F\left(w_G^{k+1}\right)-F\left(w_G^k\right) \leq \\
&\sum_{m=1}^M \mathds{1}\{k+1 \bmod m=0\} \alpha_m^{k+1}\left[F\left(w_{I, m}^{k+1}\right)-F\left(w_G^k\right)\right].
\end{aligned}
\end{equation}}

Then, we take expectations on both sides, we have 
{\small
\begin{equation}
\begin{aligned}
     &\mathbb{E}\left[F\left(w_G^{k+1}\right)\right]-\mathbb{E}\left[F\left(w_G^k\right)\right] \leq \\
     &\mathbb{E}\left\{\sum_{m=1}^M \mathds{1}\{k+1 \bmod m=0\} \alpha_m^{k+1}\left[F\left(w_{I, m}^{k+1}\right)-F\left(w_G^k\right)\right]\right\}.
\end{aligned}
\end{equation}
}

According to the theorem and lemma from \cite{zhou2022time}, we get 
{\small
\begin{equation}
   \begin{aligned}
    F(w_{I,m}^{k+1}) - & F(w^k_G) \leq \\
        &(w_{I,m}^{k+1} - w^k_G)^T \nabla F(w^k_G) + \frac{L}{2} \|w_{I,m}^{k+1} - w^k_G\|^2 .
   \end{aligned}
\end{equation}}

Since $w^{k+1}_{G} = w^{k+1-m}_G - \lambda k  $ \quad  and \quad  $ k = \sum_{u \in S_m} \frac{\nabla F(w^{k+1}_{G})}{D_m} $ \quad  with $\lambda$ as the learning ratio, we get 
{\small 
\begin{equation}
 \begin{aligned}
    &F(w^{k+1}_{I,m}) - F(w^k_G)  \leq  \\
    &(w^{k+1-m}_G - w^k_G - \lambda k)^T \nabla F(w^k_G) + \frac{L}{2} \|w_G^{k+1-m} - w^k_G - \lambda k \|^2 .\\
\end{aligned}
\end{equation}}

Because \( \|x - y\|^2 \leq 2(\|x\|^2 + \|y\|^2) \), we set \( \lambda = \frac{1}{2L} \) and utilize assumption 2, Eq. (\ref{eq17}\&\ref{eq18}). Consequently, by incrementing \( k \) in the assumption, we conclude 

{\small
\begin{equation}
\begin{aligned}
    F(w_{I,m}^{k+1})& - F(w^k_G) \leq  \\
    & L\varepsilon^2 + (\delta - \frac{1}{4L}) \|\nabla F(w^k_G)\|^2 + \frac{1}{4L} \|\nabla F(w^k_G) - k\|^2.
\end{aligned}
\end{equation}}

Then we have  
{\small
\begin{equation}
\begin{aligned}
&\mathbb{E}\left[F\left(\omega_G^{k+1}\right)\right]-\mathbb{E}\left[F\left(\omega_G^k\right)\right] \leqslant \\
&\mathbb{E}\left\{\sum_{m=1}^M \mathds{1}\{k+1 \bmod m=0\} \alpha_m^{k+1} \bigg[ L \varepsilon^2 + \right.\bigg.\\
&\bigg.\left.\left(\delta-\frac{1}{4 L}\right)\left\|\nabla F\left(\omega_G^k\right)\right\|^2+\frac{1}{4 L} \| \nabla F\left(\omega_G^k\right)- k \|^2 \right] \bigg\}.
\end{aligned}
\end{equation}}

For the right-hand side, applying the mean value theorem, we find a positive number \( \xi \in (0, M) \). 

Thus, we have
{\small
\begin{equation}
\begin{aligned}
    \mathbb{E}[&F(w_G^{k+1})]  - \mathbb{E}[F(w_G^k)] \leqslant \\
     \xi\frac{1}{M} & \sum_{m=1}^M \left( L \varepsilon^2 - \left(\frac{1}{4L} - \delta\right) \mathbb{E}\|\nabla F(w_{G}^k)\|^2 + \frac{1}{4L} \mathbb{E}\|\nabla F(w_G^k) - k\|^2\right) .
     \label{eq36}
\end{aligned} 
\end{equation}}

Since
{\small
\begin{subequations}\label{eq37}
\begin{align}
k &= \sum_{u \in S_m} \frac{\nabla f\left(w_G^{k+1-m}; u\right)}{D_m} ,
    \label{eq37a} \\[4pt]
\nabla F\left(w_G^k\right) &= \sum_{j=1}^M \sum_{u \in S_j} \frac{\nabla f\left(w_G^{k};u\right)}{D} 
    \label{eq37b}
\end{align}
\end{subequations}
}

and $D_m$ is the number of data sample in $m$-tier, we have
{\small
\begin{equation}
\begin{aligned}
& \mathbb{E}\|\nabla F(w_G^k) - k\|^2  \leq 
\\ 
& \mathbb{E}\left\| \Omega + \sum_{u \in S_m} \frac{\nabla f(w_G^k; u)}{D_m} - \sum_{u \in S_m} \frac{\nabla f(w_G^{k+1-m}; u)}{D_m} \right\|^2 .
\end{aligned}
\end{equation}}

Based on triangle inequality: $\quad \|a+b\|^2 \leqslant(\|a\|+\|b\|)^2 \quad $ and $ \Omega  = \sum_{j=1, j \neq m}^M \sum_{u \in S_j} \frac{\nabla f(w_{G,u}^{k,m})}{D} $,  we have 
{\small
\begin{equation}
\begin{aligned}
&\mathbb{E}\|\nabla F(w_G^k) - k\|^2 \leq \\
&\mathbb{E}\left(\|\Omega\|+\left\|\sum_{u \in S_m} \frac{\nabla f(w_G^k;u)- \nabla f(w_G^{k+1-m};u)}{D_m}\right\|\right)^2 .
\end{aligned}
\end{equation}}

Based on \( L \)-smoothness, and by defining \( w_G^{k, 0} \) as the pruned \( w_G^k \) where \( w_G^{k, 0} = w_G^k \odot m_G^k \), we consequently have 
{\small\begin{equation}
\begin{aligned}
&\mathbb{E}\|\nabla F(w_G^k) - k\|^2 \\
&\leq \mathbb{E}\bigg(\|\Omega\|+\bigg\|\sum_{u \in S_m} L\left(w_{G, m}^{k, u}-w_{G, m}^{k+1-m, u}\right) / D_m\bigg\|\bigg)^2  \\
&= \mathbb{E}\bigg(\|\Omega\|+\bigg\|\sum_{u \in S_m} L\left(w_{G, m}^{k, u}-w_{G, m}^{k, u, 0}+ \right. \bigg. \bigg.\\
& \quad \bigg.\left. w_{G, m}^{k, u, 0}-w_{G, m}^{k, u}+w_{G, m}^{k, u}-w_{G, m}^{k+1-m, u}\right) / D_m\bigg\|\bigg)^2 .
\end{aligned}
\end{equation}}

Based on Cauchy-Schwartz inequality: $(x+y)^2 \leqslant 2\left(x^2+y^2\right)$ and Jensen's inequality, we have 
{\small
\begin{equation}
\begin{aligned}
&\mathbb{E}\|\nabla F(w_G^k) - k\|^2 \\
&\leq 2 \mathbb{E}\|\Omega\|^2+2 L^2 S_m \sum_{u \in S m} \frac{1}{D_m^2} \mathbb{E}\left\| (w_{G, m}^k-w_{G, m}^{k, 0})+ \right.\\
&\left.(w_{G, m}^{k, 0}-w_{G, m}^k)+(w_{G, m}^k-w_{G, m}^{k+1-m})\right\|^2
\end{aligned}
\end{equation}}

Based on $(x+y+z)^2 \leqslant 3\left(x^2+y^2+z^2\right)$, considering pruned noise $\mathbb{E}\left\|w_{G,m}^k-w_{G,m}^{k, 0}\right\|^2 \leqslant \rho_{k, m} D^2$, $\|a+b\| \leqslant\|a\|+\|b\|$. 

Denote $S_m$ as the number of users in $m$-item, and the bounded gradient $\mathbb{E}\left\|\nabla F_n\left(W_{G, u}^{k, m}, \xi_{G, u}^{k, m}\right)\right\|^2 \leqslant \phi^2$, we conclude 
\vspace{-0.2cm}
{\small\begin{equation} \begin{aligned}
&\mathbb{E}\|\nabla F(w_G^k) - k\|^2\\
&\leq 2\mathbb{E}\left\|\sum_{j=1,j\neq m}^M \sum_{u\in S_j} \frac{\nabla f(w_{G,u}^{k,m})}{D} \right\|^2 \\
& +6 L^2 S_m \sum_{u \in S m} \frac{1}{D_m^2} \left(\underbrace{\mathbb{E}\|w_{G, m}^k-w_{G, m}^{k, 0}\|^2}_{ \rho_{k, m} D^2}+ \right.\\
& \left. \underbrace{\mathbb{E}\|w_{G, m}^{k, 0}-w_{G, m}^k\|^2}_{\rho_{k, m} D^2}+\underbrace{\mathbb{E}\|w_{G, m}^k-w_{G, m}^{k+1-m}\|^2}_{\varepsilon^2}\right).
\end{aligned} 
\end{equation}}
\vspace{-0.6cm}
Then, we have 
\vspace{-0.6cm}
{\small\begin{equation}
\begin{aligned}
\mathbb{E}\|\nabla F(w_G^k) - k\|^2 &\leq 2 \frac{M-1}{D^2} \sum_{j=1, j \neq m}^M S_j \sum_{u \in S_j} \phi^2+ \\
&6 S_m \sum_{u \in S_m}\left(\frac{L}{D_m}\right)^2\left(2 \rho_{k, m} D^2+\varepsilon^2\right).
\end{aligned} 
\end{equation}}

Therefore, the right-hand side item of Eq. (\ref{eq36}) can be  
{\small\begin{equation}
\begin{aligned}
&\mathbb{E}\left[F\left(w_G^{k+1}\right)\right]-\mathbb{E}\left[F\left(w_G^k\right)\right]  \\
&\leq \xi \frac{1}{M} \sum_{m=1}^M\left[L \varepsilon^2-\left(\frac{1}{4 L}-\delta\right) \mathbb{E}\left\|\nabla F\left(w_G^k\right)\right\|^2 \right. \\
& \quad \quad \quad\quad\quad\quad\quad\quad\quad \left. + \frac{1}{4 L} \mathbb{E}\left\|\nabla F\left(w_G^k\right)-k\right\|^2\right] \\
&\leq \xi \frac{1}{M} \sum_{m=1}^M\left\{L \varepsilon^2-\left(\frac{1}{4 L}-\delta\right) \mathbb{E}\left\|\nabla F\left(w_G^k\right)\right\|^2+ \right.\\ 
&\left. \frac{M-1}{2LD^2} \sum_{j=1, j \neq m}^M S_j^2 \phi^2+S_m^2 \frac{3L}{2D_m^2}\left(2 \rho_{k,m} D^2+\varepsilon^2\right)\right\}.
\end{aligned}
\end{equation}}
\vspace{-5pt}

In the end, after changing \textit{k} to \textit{K}, we directly derive
\vspace{-2pt}
{\small
\begin{equation}
\begin{aligned}
 \quad &\mathbb{E}\left[F(w^0)\right] - \mathbb{E}\left[F(w^*)\right]\\
& = \sum_{k=1}^K \mathbb{E}\left[F(w_G^{k+1})\right] - \sum_{k=1}^K \mathbb{E}\left[F(w_G^k)\right] \\
&\leq \sum_{k=1}^K \xi \frac{1}{M} \sum_{m=1}^M L \varepsilon^2 - \sum_{k=1}^K \xi \frac{1}{M} \sum_{m=1}^M \left(\frac{1}{4L} - \delta\right) \mathbb{E}\left\|\nabla F(\omega_G^k)\right\|^2 \\
&+\xi K \frac{M-1}{2 L M D^2} \sum_{m=1}^M  \sum_{j=1, j \neq m}^M S_j^2 \phi^2+ \\
&\xi \frac{1}{M} \sum_{k=1}^K \sum_{m=1}^M S_m^2 \frac{3 L}{2 D_m^2}\left(2 \rho_{k, m} D^2+\varepsilon^2\right),
\end{aligned}
\end{equation}}
where $w^0$ is the initial global model. Therefore, when $\frac{1}{4 L}-\delta>0$ holds, the upper bound of gradients is derived as
\vspace{-5pt}
{\small
\begin{equation}
\begin{aligned}
\xi &\sum_{k=1}^K \left(\frac{1}{4 L}-\delta\right) \mathbb{E}\left\|\nabla F\left(w_G^k\right)\right\|^2  \\
&\leq \mathbb{E}\left[F\left(w^0\right)\right]-\mathbb{E}\left[F\left(w^*\right)\right] +K \xi L \varepsilon^2 + \\
&\xi \frac{1}{M} \sum_{k=1}^K \sum_{m=1}^M S_m^2 \frac{3L}{2D_m^2}\left(2 \rho_{k, m} D^2+\varepsilon^2\right)+ \\
&\xi K \frac{M-1}{2 L M D^2}  \sum_{m=1}^M \sum_{j=1, j \neq m}^M S_j^2 \phi^2 \\
& = \mathbb{E}\left[F\left(w^0\right)\right]-\mathbb{E}\left[F\left(w^*\right)\right] \\
&+K \xi L \varepsilon^2 + \xi \frac{3LD^2}{M} \sum_{k=1}^K \sum_{m=1}^M \frac{S_m^2}{D_m^2}  \rho_{k, m} + \\
&\xi \frac{3KL}{2M} \sum_{m=1}^M  \frac{S_m^2}{D_m^2} \varepsilon^2+\xi K \frac{M-1}{2 L M D^2}  \sum_{m=1}^M \sum_{j=1, j \neq m}^M S_j^2 \phi^2 ,
\end{aligned}
\end{equation}
}
which ends the proof of Theorem 1.

\section{Proof of the Theorem 2 }
According to the last line of Eq. (\ref{eq24}), pruning ratio $\rho^*_{m, k}$ is calculated as
{\small
\begin{subequations}\label{eq47}
\begin{gather}
\left(1-\rho_{k, m}\right) W_{m, k}^u\left(\zeta \frac{c_u}{f_u}+\frac{\hat{q}}{R_{m, u}^k}\right) 
\leq m \Delta T 
\tag{\color{black}{47a}} \label{eq47a} \\[6pt]
\left(  W_{u, \text{conv}}  + \left(1-\rho_{k, m}\right) W_{u, \text{fully}} \right)
\left(\zeta \frac{c_u}{f_u}+\frac{\hat{q}}{R_{m, u}^k}\right) 
\leq m \Delta T 
\tag{\color{black}{47b}} \label{eq47b} \\[6pt]
\left(1-\rho_{k, m}\right)  \leq 
\frac{ m \Delta T - W_{u, \text{conv}}\left(\zeta \frac{c_u}{f_u}+\frac{\hat{q}}{R_{m, u}^k}\right)}
{ W_{u, \text{fully}} \left(\zeta \frac{c_u}{f_u}+\frac{\hat{q}}{R_{m, u}^k}\right) }
\tag{\color{black}{47c}} \label{eq47c}
\end{gather}
\end{subequations}
}

Therefore, the $\rho^*_{m, k}$ is deduced as Eq. (\ref{eq25}), which ends the proof.
\section{Proof of the Lemma 1 }
The objective function in Eq. (\ref{eq26}) is equivalent to
{\small
\begin{equation}
    F(X)=\sum_{k=1}^K \sum_{m=1}^M f\left(x_{k,m} \right)=\sum_{k=1}^K \sum_{m=1}^M\left(1-\frac{V_1-\frac{V_2}{x_{k,m}}}{V_3+\frac{V_4}{x_{k,m}}}\right),
\end{equation}}
where $V_1, V_2, V_3, V_4 >0$, which means the constant value and $0<x_{k,m}<1$. 

To prove the lemma, we should first analyze the convexity of the function $f\left(x_{k,m}\right)$. The second derivative can be derived as
{\small
\begin{equation}
    f^{\prime \prime}\left(x_{k,m}\right)=\frac{2 V_3\left(V_1 V_4+V_2 V_3\right)\left(V_3 x_{k,m}+V_4\right)}{\left(V_3 x_{k,m}+V_4\right)^4} \geq 0 .
\end{equation}}

Based on the theorem \cite{boyd2004convex}, the objective function in Eq. (\ref{eq26}) is convex. In addition, both constraints Eq. (\ref{eq15c}) and Eq. (\ref{eq15d}) are also convex. 

Therefore, the optimization problem in Eq. (\ref{eq26}) is convex, which ends the proof of Lemma 1.

\section{Proof of the Theorem 3 }
According to the problem in Eq. (\ref{eq23}) and Eq. (\ref{eq15}) , the Lagrange function can be constructed as 
{\small
\begin{equation}
\resizebox{\columnwidth}{!}{$
\begin{aligned}
    \mathcal{L}\left(b_{m,k}, \lambda\right)
    &= \boldsymbol{\Delta} \sum_{k=1}^K \sum_{m=1}^M \frac{S_m^2}{D_m^2}
    \left(1-\frac{m \Delta T - W_{u, \text{conv}}
        \left(\xi \frac{ c_u}{f_u}+\frac{\hat{q}}{R_{m,u}^{\text{k}}}\right)}
        {(\xi \frac{ c_u}{f_u}+\frac{\hat{q}}{R_{m,u}^{\text{k}}})W_{u, \text{fully}}}\right)
    \\
    &\quad + \lambda\left(\sum_{m=1}^M b_{m,k} -1\right)
\end{aligned}
$}
\end{equation}
}

Where $\lambda$ is the Lagrange multiplier. Then, the Karush-KuhnTucker (KKT) conditions can be derived as
{\small
\begin{subequations}\label{eq51}
\begin{gather}
\frac{\partial \mathcal{L}}{\partial b_{m,k}} =
\lambda - \frac{
\boldsymbol{\Delta} S_m^2\, m \Delta T\, \hat{q} W_{u,\text{fully}} B \log_2\left(1+\frac{p g_{k,u}}{\sigma^2}\right)
}{
D_m^2 \left(b_{m,k} B W_{u,\text{fully}} \xi \dfrac{c_u}{f_u} \log_2\left(1+\dfrac{p g_{k,u}}{\sigma^2}\right)
+ \hat{q} W_{u,\text{fully}}\right)^2
}
\notag \\[2pt]
= 0,\quad m \in \mathcal{M}
\tag{\color{black}{51a}} \label{eq51a} \\[6pt]
\lambda\left(\sum_{m=1}^M b_{m,k} -1\right) = 0
\tag{\color{black}{51b}} \label{eq51b} \\[6pt]
\lambda \geq 0
\tag{\color{black}{51c}} \label{eq51c}
\end{gather}
\end{subequations}
}

Based on these KKT conditions, the optimal wireless bandwidth allocation can be obtained as Theorem 3, which ends the proof.



\vspace{-7pt}
\ifCLASSOPTIONcaptionsoff
  \newpage
\fi




\vspace{5pt}
\small
\bibliographystyle{IEEEtran}
\bibliography{main}

@ARTICLE{9144301,
  author={Chowdhury, Mostafa Zaman and Shahjalal, Md. and Ahmed, Shakil and Jang, Yeong Min},
  journal={IEEE Open J. Commun. Soc.}, 
  title={6G Wireless Communication Systems: Applications, Requirements, Technologies, Challenges, and Research Directions}, 
  year={Jul. 2020},
  volume={1},
  number={},
  pages={957-975},
  doi={10.1109/OJCOMS.2020.3010270}}

@article{sun2014data,
  title={Data security and privacy in cloud computing},
  author={Sun, Yunchuan and Zhang, Junsheng and Xiong, Yongping and Zhu, Guangyu},
  journal={Int. J. Distrib. Sens. Netw.},
  volume={10},
  number={7},
  pages={190903},
  year={Jul. 2014},
  publisher={SAGE Publications Sage UK: London, England}
}

@article{zhang2021survey,
  title={A survey on federated learning},
  author={Zhang, Chen and Xie, Yu and Bai, Hang and Yu, Bin and Li, Weihong and Gao, Yuan},
  journal={Knowledge-Based Systems},
  volume={216},
  pages={106775},
  year={Mar. 2021},
  publisher={Elsevier}
}

@inproceedings{mcmahan2017communication,
  title={Communication-efficient learning of deep networks from decentralized data},
  author={McMahan, Brendan and Moore, Eider and Ramage, Daniel and Hampson, Seth and y Arcas, Blaise Aguera},
  booktitle={Artificial intelligence and statistics},
  pages={1273--1282},
  year={Apr. 2017},
  organization={PMLR}
}

@article{xu2023asynchronous,
  title={Asynchronous federated learning on heterogeneous devices: A survey},
  author={Xu, Chenhao and Qu, Youyang and Xiang, Yong and Gao, Longxiang},
  journal={Comput. Sci. Rev.},
  volume={50},
  pages={100595},
  year={Nov. 2023},
  publisher={Elsevier}
}

@article{chai2020fedat,
  title={Fedat: A communication-efficient federated learning method with asynchronous tiers under non-iid data},
  author={Chai, Zheng and Chen, Yujing and Zhao, Liang and Cheng, Yue and Rangwala, Huzefa},
  journal={ArXivorg},
  year={Oct. 2020}
}

@article{wang2019adaptive,
  title={Adaptive federated learning in resource constrained edge computing systems},
  author={Wang, Shiqiang and Tuor, Tiffany and Salonidis, Theodoros and Leung, Kin K and Makaya, Christian and He, Ting and Chan, Kevin},
  journal={IEEE J. Sel. Areas Commun.},
  volume={37},
  number={6},
  pages={1205--1221},
  year={Mar. 2019},
  publisher={IEEE}
}

@article{yang2020federated,
  title={Federated learning via over-the-air computation},
  author={Yang, Kai and Jiang, Tao and Shi, Yuanming and Ding, Zhi},
  journal={IEEE Trans. Wireless Commun.},
  volume={19},
  number={3},
  pages={2022--2035},
  year={Jan. 2020},
  publisher={IEEE}
}

@article{yang2020energy,
  title={Energy efficient federated learning over wireless communication networks},
  author={Yang, Zhaohui and Chen, Mingzhe and Saad, Walid and Hong, Choong Seon and Shikh-Bahaei, Mohammad},
  journal={IEEE Trans. Wireless Commun.},
  volume={20},
  number={3},
  pages={1935--1949},
  year={Mar. 2020},
  publisher={IEEE}
}

@inproceedings{tran2019federated,
  title={Federated learning over wireless networks: Optimization model design and analysis},
  author={Tran, Nguyen H and Bao, Wei and Zomaya, Albert and Nguyen, Minh NH and Hong, Choong Seon},
  booktitle={Proc. IEEE Conf. Comput. Commun. (INFOCOM)},
  pages={1387--1395},
  year={Jun. 2019},
  organization={IEEE}
}

@article{jiang2022model,
  title={Model pruning enables efficient federated learning on edge devices},
  author={Jiang, Yuang and Wang, Shiqiang and Valls, Victor and Ko, Bong Jun and Lee, Wei-Han and Leung, Kin K and Tassiulas, Leandros},
  journal={IEEE Trans. Neural Netw. Learn Syst.},
  year={Dec. 2022},
  publisher={IEEE}
}

@article{liu2023adaptive,
  title={Adaptive Federated Pruning in Hierarchical Wireless Networks},
  author={Liu, Xiaonan and Wang, Shiqiang and Deng, Yansha and Nallanathan, Arumugam},
  journal={IEEE Trans. Wireless Commun.},
  year={Nov. 2023},
  publisher={IEEE}
}

@article{xie2019asynchronous,
  title={Asynchronous federated optimization},
  author={Xie, Cong and Koyejo, Sanmi and Gupta, Indranil},
  journal={arXiv preprint arXiv:1903.03934},
  year={2019}
}

@article{konevcny2016federated,
  title={Federated learning: Strategies for improving communication efficiency},
  author={Kone{\v{c}}n{\`y}, Jakub and McMahan, H Brendan and Yu, Felix X and Richt{\'a}rik, Peter and Suresh, Ananda Theertha and Bacon, Dave},
  journal={arXiv preprint arXiv:1610.05492},
  year={2016}
}

@article{chai2021fedat,
  title={FedAT: A high-performance and communication-efficient federated learning system with asynchronous tiers},
  author={Chai, Z. and Chen, Y. and Zhao, L. and Cheng, Y. and Rangwala, H.},
  journal={CoRR},
  volume={abs/2010.05958},
  pages={1--16},
  month={Oct. },
  year={2020}
}

@inproceedings{reisizadeh2020fedpaq,
  title={Fedpaq: A communication-efficient federated learning method with periodic averaging and quantization},
  author={Reisizadeh, Amirhossein and Mokhtari, Aryan and Hassani, Hamed and Jadbabaie, Ali and Pedarsani, Ramtin},
  booktitle={Proc. Int. Conf.
Artif. Intell. Stat},
  pages={2021--2031},
  year={Aug. 2020},
}

@article{zhou2022time,
  title={Time-triggered federated learning over wireless networks},
  author={Zhou, Xiaokang and Deng, Yansha and Xia, Huiyun and Wu, Shaochuan and Bennis, Mehdi},
  journal={IEEE Trans. Wireless Commun.},
  volume={21},
  number={12},
  pages={11066-11079},
  year={Dec. 2022},
  publisher={IEEE}
}

@InProceedings{cho2020client,
  title = 	 { Towards Understanding Biased Client Selection in Federated Learning },
  author =       {Jee Cho, Yae and Wang, Jianyu and Joshi, Gauri},
  booktitle = 	 {Proc. Int. Conf. Artif. Intell. Statist.
(AISTATS)},
  pages = 	 {10351--10375},
  year = 	 {2022},
  editor = 	 {Camps-Valls, Gustau and Ruiz, Francisco J. R. and Valera, Isabel},
  volume = 	 {151},
  month = 	 {Mar. },
 }

@inproceedings{zhou2022resource,
  title={Resource allocation for time-triggered federated learning over wireless networks},
  author={Zhou, Xiaokang and Deng, Yansha and Xia, Huiyun and Wu, Shaochuan and Bennis, Mehdi},
  booktitle={Proc. IEEE Int. Conf. Commun.},
  pages={1-30},
  year={Apr. 2022},
}

@inproceedings{molchanov2019importance,
  title={Importance estimation for neural network pruning},
  author={Molchanov, Pavlo and Mallya, Arun and Tyree, Stephen and Frosio, Iuri and Kautz, Jan},
  booktitle={Proc. IEEE/CVF Conf. Comput. Vis. Pattern Recognit. (CVPR)},
  pages={11264--11272},
  year={Jun. 2019}
}

@article{ghadimi2013stochastic,
  title={Stochastic first-and zeroth-order methods for nonconvex stochastic programming},
  author={Ghadimi, Saeed and Lan, Guanghui},
  journal={SIAM J. Optim.},
  volume={23},
  number={4},
  pages={2341--2368},
  year={Dec. 2013},
  publisher={SIAM}
}

@inproceedings{shi2019convergence,
  title={A convergence analysis of distributed SGD with communication-efficient gradient sparsification.},
  author={Shi, Shaohuai and Zhao, Kaiyong and Wang, Qiang and Tang, Zhenheng and Chu, Xiaowen},
  booktitle={IJCAI},
  pages={3411--3417},
  year={Aug. 2019}
}

@inproceedings{stich2018sparsified,
  title={Sparsified SGD with memory},
  author={Stich, Sebastian U and Cordonnier, Jean-Baptiste and Jaggi, Martin},
  journal={Proc. Adv. Neural Inf. Process. Syst. (NeurIPS’18)},
  volume={31},
  year={Dec. 2018}
}

@article{dinh2020federated,
  title={Federated learning over wireless networks: Convergence analysis and resource allocation},
  author={Dinh, Canh T and Tran, Nguyen H and Nguyen, Minh NH and Hong, Choong Seon and Bao, Wei and Zomaya, Albert Y and Gramoli, Vincent},
  journal={IEEE/ACM Trans. Netw.},
  volume={29},
  number={1},
  pages={398--409},
  year={Nov. 2020},
  publisher={IEEE}
}

@article{belotti2013mixed,
  title={Mixed-integer nonlinear optimization},
  author={Belotti, Pietro and Kirches, Christian and Leyffer, Sven and Linderoth, Jeff and Luedtke, James and Mahajan, Ashutosh},
  journal={Acta Numerica},
  volume={22},
  pages={1--131},
  year={Apr. 2013},
  publisher={Cambridge University Press}
}

@incollection{bomze2010interior,
  title={Interior point methods for nonlinear optimization},
  author={Pólik, I. and Terlaky, T.},
  booktitle={Nonlinear Optimization},
  publisher={Springer},
  address={Berlin, Germany},
  month={Jan.},
  year={2010},
  pages={215--276}
}

@article{lecun1998gradient,
  title={Gradient-based learning applied to document recognition},
  author={LeCun, Yann and Bottou, L{\'e}on and Bengio, Yoshua and Haffner, Patrick},
  journal={Proc. IEEE},
  volume={86},
  number={11},
  pages={2278--2324},
  year={Nov. 1998},
  publisher={Ieee}
}

@book{boyd2004convex,
  title={Convex optimization},
  author={Boyd, Stephen P and Vandenberghe, Lieven},
  year={2004},
  publisher={Cambridge University Press}
}

@article{hassibi1992second,
  title={Second order derivatives for network pruning: Optimal brain surgeon},
  author={Hassibi, Babak and Stork, David},
  journal={Adv. Neural Inf. Process. Syst.},
  volume={5},
  year={1992}
}

@article{liu2019channel,
  title={Channel pruning based on mean gradient for accelerating convolutional neural networks},
  author={Liu, Congcong and Wu, Huaming},
  journal={Signal Processing},
  volume={156},
  pages={84--91},
  year={2019},
  publisher={Elsevier}
}

@article{lee2018snip,
  title={Snip: Single-shot network pruning based on connection sensitivity},
  author={Lee, Namhoon and Ajanthan, Thalaiyasingam and Torr, Philip HS},
  journal={arXiv preprint arXiv:1810.02340},
  year={2018}
}

@article{castellano1997iterative,
  title={An iterative pruning algorithm for feedforward neural networks},
  author={Castellano, Giovanna and Fanelli, Anna Maria and Pelillo, Marcello},
  journal={IEEE Trans. Neural Netw.},
  volume={8},
  number={3},
  pages={519--531},
  year={1997},
  publisher={IEEE}
}

@article{tanaka2020pruning,
  title={Pruning neural networks without any data by iteratively conserving synaptic flow},
  author={Tanaka, Hidenori and Kunin, Daniel and Yamins, Daniel L and Ganguli, Surya},
  journal={Adv. Neural Inf. Process. Syst.},
  volume={33},
  pages={6377--6389},
  year={2020}
}

@article{wen2022federated,
  title={Federated dropout—A simple approach for enabling federated learning on resource constrained devices},
  author={Wen, Dingzhu and Jeon, Ki-Jun and Huang, Kaibin},
  journal={IEEE Wirel. Commun. Lett.},
  volume={11},
  number={5},
  pages={923--927},
  year={2022},
  publisher={IEEE}
}

@article{chen2020joint,
  title={A joint learning and communications framework for federated learning over wireless networks},
  author={Chen, Mingzhe and Yang, Zhaohui and Saad, Walid and Yin, Changchuan and Poor, H Vincent and Cui, Shuguang},
  journal={IEEE Trans. Wireless Commun.},
  volume={20},
  number={1},
  pages={269--283},
  year={2020},
  publisher={IEEE}
}

@article{shi2020joint,
  title={Joint device scheduling and resource allocation for latency constrained wireless federated learning},
  author={Shi, Wenqi and Zhou, Sheng and Niu, Zhisheng and Jiang, Miao and Geng, Lu},
  journal={IEEE Trans. Wireless Commun.},
  volume={20},
  number={1},
  pages={453--467},
  year={2020},
  publisher={IEEE}
}

@inproceedings{zhang2024joint,
  title={Joint Model Pruning and Resource Allocation for Wireless Time-triggered Federated Learning},
  author={Zhang, Xinlu and Deng, Yansha and Mahmoodi, Toktam},
  booktitle={GLOBECOM 2024-2024 IEEE Global Communications Conference},
  pages={950--955},
  year={2024},
  organization={IEEE}
}

@ARTICLE{10660465,
  author={Ding, Yahao and Shang, Wen and Yang, Yinchao and Ding, Weihang and Shikh-Bahaei, Mohammad},
  journal={IEEE Internet of Things Journal}, 
  title={Joint Layer Selection and Differential Privacy Design for Federated Learning Over Wireless Networks}, 
  year={2024},
  volume={11},
  number={24},
  pages={39767-39779},
  doi={10.1109/JIOT.2024.3452549}}

@ARTICLE{10353003,
  author={Ding, Yahao and Yang, Zhaohui and Pham, Quoc-Viet and Hu, Ye and Zhang, Zhaoyang and Shikh-Bahaei, Mohammad},
  journal={IEEE Internet of Things Journal}, 
  title={Distributed Machine Learning for UAV Swarms: Computing, Sensing, and Semantics}, 
  year={2024},
  volume={11},
  number={5},
  pages={7447-7473},
  doi={10.1109/JIOT.2023.3341307}}

\vfill

\end{document}